\documentclass[runningheads]{llncs}

\usepackage{eccv}

\usepackage{eccvabbrv}

\usepackage{graphicx}
\usepackage{booktabs}
\usepackage{amsmath}
\usepackage{amssymb}
\usepackage{multirow}
\usepackage{array}
\usepackage{makecell}
\usepackage{amsfonts}
\usepackage{algorithm}
\usepackage{algpseudocode}
\usepackage{setspace}
\usepackage{enumitem}
\usepackage{bbm}
\usepackage{bm}
\usepackage{xcolor}
\usepackage{pifont}
\usepackage{courier}
\usepackage{wrapfig}

\usepackage{cuted}
\usepackage{capt-of}

\newcommand{\cmark}{\text{\ding{51}}}
\newcommand{\xmark}{\text{\ding{55}}}

\usepackage{colortbl}

\definecolor{c_muns}{rgb}{0.88,1,1}
\definecolor{c_msup}{rgb}{1.0,0.88,1}
\definecolor{c_data}{rgb}{0.9,0.9,0.9}
\definecolor{c_lowbest}{rgb}{1.0,0.8,0.8}
\definecolor{c_highbest}{rgb}{0.8,0.8,1.0}

\usepackage[accsupp]{axessibility}  

\usepackage{hyperref}

\usepackage{orcidlink}

\begin{document}

\title{SkateFormer: Skeletal-Temporal Transformer\\for Human Action Recognition} 

\titlerunning{SkateFormer: Skeletal-Temporal Transformer}

\author{Jeonghyeok Do\orcidlink{0000-0003-0030-0129} \and
Munchurl Kim\orcidlink{0000-0003-0146-5419}}

\authorrunning{J. Do and M. Kim}

\institute{Korea Advanced Institute of Science and Technology, South Korea \newline
\email{\{ehwjdgur0913, mkimee\}@kaist.ac.kr} \newline
\url{https://kaist-viclab.github.io/SkateFormer_site/}}
\maketitle

\begin{abstract}
Skeleton-based action recognition, which classifies human actions based on the coordinates of joints and their connectivity within skeleton data, is widely utilized in various scenarios. While Graph Convolutional Networks (GCNs) have been proposed for skeleton data represented as graphs, they suffer from limited receptive fields constrained by joint connectivity. To address this limitation, recent advancements have introduced transformer-based methods. However, capturing correlations between all joints in all frames requires substantial memory resources. To alleviate this, we propose a novel approach called Skeletal-Temporal Transformer (SkateFormer) that partitions joints and frames based on different types of skeletal-temporal relation (Skate-Type) and performs skeletal-temporal self-attention (Skate-MSA) within each partition. We categorize the key skeletal-temporal relations for action recognition into a total of four distinct types. These types combine (i) two skeletal relation types based on physically neighboring and distant joints, and (ii) two temporal relation types based on neighboring and distant frames. Through this partition-specific attention strategy, our SkateFormer can selectively focus on key joints and frames crucial for action recognition in an action-adaptive manner with efficient computation. Extensive experiments on various benchmark datasets validate that our SkateFormer outperforms recent state-of-the-art methods. 
\end{abstract}

\begin{sloppypar}

\section{Introduction}
\label{sec:intro}

In recent years, human action recognition (HAR) \cite{Survey1, Survey2, Survey3} has gained widespread applications in real-life scenarios, involving the classification of actions based on human movements. A diverse range of data sources such as videos captured from RGB cameras, optical flow generated through post-processing, 2D/3D skeletons estimated from RGB videos \cite{HRNet, Openpose}, and skeletons acquired from sensors \cite{Kinetics}, contain information about human movements that can be leveraged for action recognition. However, using RGB videos as input is challenging. They are sensitive to external factors such as lighting conditions, camera distance and angles, and background variations \cite{PoseC3D}. Moreover, RGB videos require large storage capacity due to their detailed background visuals. Conversely, 3D skeletons obtained through sensors can offer a compact representation with robustness to external environmental changes, and do not require additional post-processing modules such as pose and optical flow estimations \cite{Flownet, PWC, Openpose, HRNet}.

\begin{figure}[tbp]
  \centering
  \includegraphics[width=\textwidth]{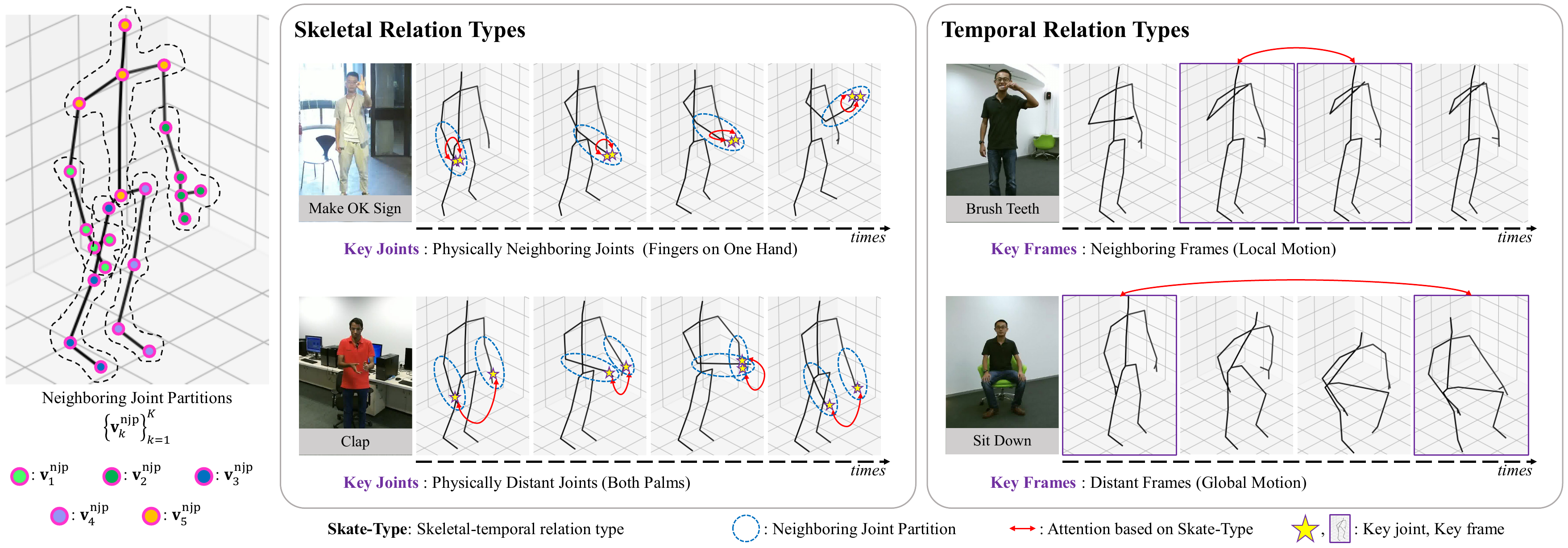}
  \captionof{figure}{SkateFormer's partition-specific attention strategy for skeleton-based action recognition, leveraging two skeletal relation types -- physically neighboring and distant joints -- and two temporal relation types for local and global motion.}
  \label{fig:1}
  \vspace{-0.6cm}
\end{figure}

The joints and their connections (or bones) in skeleton input correspond to the vertices and edges, respectively, in a graph structure. Consequently, many methods based on Graph Convolutional Networks (GCNs) \cite{GCN} suitable for processing graph inputs have been extensively proposed for human action recognition. Most of the GCN-based methods \cite{AGCN, ASGCN, DCGCN, STGCN, MSTGCN, NASGCN, STGCN++, ShiftGCN, EfficientGCN, DDGCN, TAGCN, PAResGCN, DynamicGCN, FRHead, LST, HDGCN, stcnet, koopman} exchange information between different joints within a single frame using graph convolutions, and capture the temporal dynamics for each joint using 1D temporal convolutions. However, they struggle to capture the relation of physically distant joints (e.g., between two hands with open arms) effectively due to the direct propagation of information between physically connected joints.

To mitigate this limitation, transformer-based methods utilizing self-attention to capture relations across all joint pairs have been proposed \cite{STSA, 3mformer, iip, mixformer}. Yet, considering every joint in every frame is inefficient as specific joints in certain frames are more critical for particular action recognition. Some transformer-based methods have tried to reduce computational complexity by squeezing features along the joint or frame dimension before self-attention \cite{skeletr}, only using either skeletal or temporal relations \cite{STST, DSTA, STTR, fgstformer, Hypergraph}, or by tokenizing physically similar skeletal information \cite{IGFormer, istanet}. Separating joints and frames during self-attention can obscure the critical skeletal-temporal relations for action recognition, and tokenization may dilute vital information, potentially degrading performance.

To overcome these issues, we propose an efficient transformer-based approach, called Skeletal-Temporal Transformer (SkateFormer) that introduces joint and frame partition strategies and partition-specific self-attention based on types of skeletal-temporal relations (Skate-Type). Fig.~\ref{fig:1} illustrates the joint and frame partition strategies and partition-specific self-attention of our SkateFormer. For example, in the `make ok sign' action class, relations between physically neighboring joints (e.g., joints on the same hand) are crucial, whereas in the `clap' action class, relations between physically distant joints (e.g., between the palms of both hands) are more critical. Regarding temporal relations, for repetitive local motions like in the `brush teeth' class, neighboring frame relations are essential, whereas for actions with global motion like `sit down', distant frame relations become critical. Additionally, the speed at which actions are performed can vary significantly depending on the actor. To make our SkateFormer computationally efficient, we introduce a novel partition-specific attention (Skate-MSA). For this, we divide the skeletal-temporal relations into four partition types: (i) neighboring joints and local motion -- Skate-Type-1, (ii) distant joints and local motion -- Skate-Type-2, (iii) neighboring joints and global motion -- Skate-Type-3 and (iv) distant joints and global motion -- Skate-Type-4. Therefore, Skate-MSA is designed to efficiently capture skeletal-temporal relations at the joint-element-level, eliminating the need for tokenization, which typically involves joint-group-level attention \cite{IGFormer, istanet}. In summary, our contributions are as follows:

\begin{itemize}
\item We propose a Skeletal-Temporal Transformer (SkateFormer), a partition-specific attention strategy (Skate-MSA) for skeleton-based action recognition that captures skeletal-temporal relations and reduces computational complexity.
\item We introduce a range of augmentation techniques and an effective positional embedding method, named Skate-Embedding, which combines skeletal and temporal features. This method significantly enhances action recognition performance by forming an outer product between learnable skeletal features and fixed temporal index features. 
\item Our SkateFormer sets a new state-of-the-art for action recognition performance across multiple modalities (4-ensemble condition) and single modalities (joint, bone, joint motion, bone motion), showing notable improvement over the most recent state-of-the-art methods. Additionally, it concurrently establishes a new state-of-the-art in interaction recognition, a sub-field of action recognition.
\end{itemize}

\section{Related Works}
\label{sec:related}

\subsection{RNN/CNN-based approaches}
Recently, skeleton-based action recognition has made significant progress. Early research efforts focused on utilizing Recurrent Neural Networks (RNNs), including LSTM and GRU, to handle skeleton data due to its sequential and continuous nature \cite{IndRNN, LSTMIRN, CoLSTM, STLSTM, VALSTM, AGCLSTM, GCALSTM}. Furthermore, some studies explored the conversion of skeleton data into pseudo-images to leverage Convolutional Neural Networks (CNNs) \cite{CNN1, CNN2, CNN3, CNN4, MTCNN, TACNN}. Very recently, PoseC3D \cite{PoseC3D} employed a 2D pose estimation network \cite{HRNet} on RGB video to generate image-like skeleton heatmaps as input for CNNs, and Ske2Grid \cite{Ske2Grid} proposed a method to transform 2D/3D skeleton data nodes into image-format grid patches.

\subsection{GCN-based approaches}
Skeleton data is composed of joints and bones, which correspond to vertices and edges in a graph. Consequently, research in this domain has predominantly gravitated towards Graph Neural Networks (GNNs) \cite{DGNN, SkeletalGNN} and Graph Convolutional Networks (GCNs) \cite{AGCN, ASGCN, DCGCN, DDGCN, TAGCN, CTRGCN, MSTGCN, NASGCN, STGCN++, PAResGCN, ShiftGCN, DynamicGCN, EfficientGCN, STGCN, stcnet, koopman}. InfoGCN  \cite{InfoGCN} introduced additional loss terms to form a compact latent feature space, resulting in clear decision boundaries. Similarly, FR-Head \cite{FRHead} incorporated contrastive learning to separate feature representations of ambiguous classes. LST \cite{LST} introduced language supervision to train GCNs effectively, while HD-GCN \cite{HDGCN} introduced rooted trees as a data structure to create diverse connections between joints. Despite their merits, GCNs have limitations in effectively capturing topology variations depending on the input data by using the fixed-sized kernels.

\subsection{Transformer-based approaches} Recent research explores transformer-based approaches for skeleton-based action recognition, which excel in capturing data-adaptive joint connections \cite{STSA, 3mformer, iip, mixformer}. However, these methods often require high computational costs due to the usage of their large-sized attention maps. Early methods have tried to reduce attention map sizes by employing feature pooling \cite{skeletr} or reshaping \cite{DSTA} techniques. Some studies have attempted to mitigate computational load by utilizing separate skeletal and temporal attention modules in a parallel \cite{STST, STTR} or serial \cite{DSTA, Hypergraph, fgstformer} configurations. Nevertheless, they face challenges in simultaneously capturing skeletal-temporal relations, which are critical aspects of skeleton-based action recognition. IGFormer \cite{IGFormer} and ISTA-Net \cite{istanet} tackle the challenge by embedding joint sets within the same partition into a unified token before attention modules, relying solely on partition strategies for tokenization without introducing partition-specific attention mechanisms. Despite incorporating skeletal-temporal attention at the joint-group-level, they encounter limitations in local action recognition due to the tokenization process leading to the loss of physically similar skeletal information. In our work, the proposed SkateFormer employs partition-specific skeletal-temporal attention modules to effectively capture skeletal-temporal relations at the joint-element-level without tokenization.

\noindent \textbf{Complexity of transformer-based vision tasks.}\quad Vision Transformers (ViT) \cite{ViT} revolutionized the field of computer vision but face challenges due to high computational costs. To mitigate these challenges, various studies have been conducted to reduce the size of attention maps while efficiently propagating local and global information \cite{Swin, Maxvit}. In the context of vision tasks, 2D feature maps are organized along horizontal and vertical spatial axes. In contrast, our task involves skeletal and temporal axes, making the nature of our data fundamentally different. However, the elements within skeleton data tend to scale mostly with the factors such as the number of recognized joints, the number of individuals performing actions, and the number of frames, necessitating the reduction of computational complexity. Inspired by the works \cite{Maxim, Maxvit}, we propose a transformer structure that efficiently captures skeletal-temporal relations at the joint-element-level, reducing the computational cost.

\section{Methodology}
\label{sec:method}

\subsection{Overview of SkateFormer}

\begin{figure}[htbp]
  \vspace{-0.5cm}
  \centering
  \includegraphics[width=0.85\linewidth]{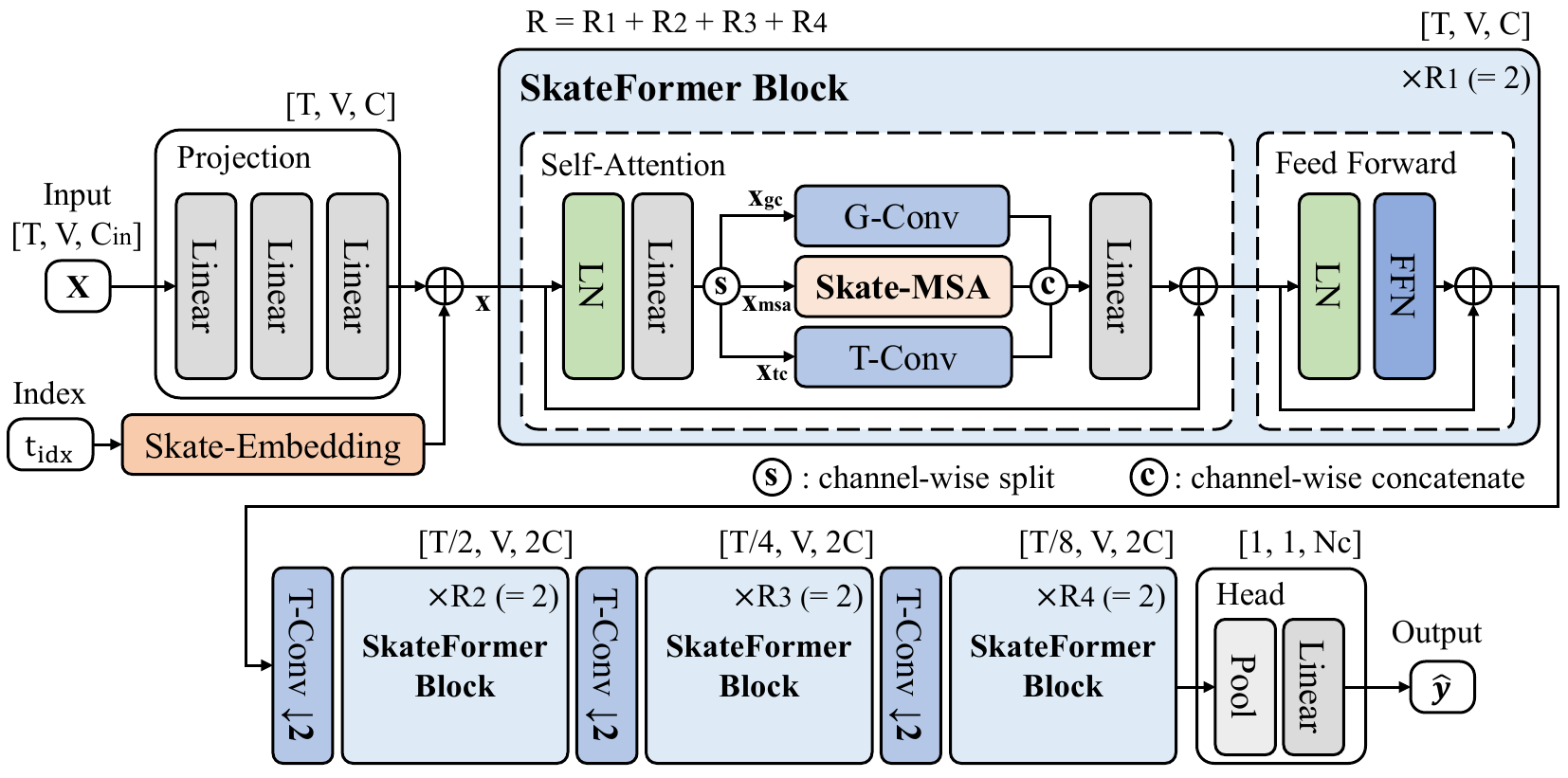}
  \caption{The overall framework of our proposed SkateFormer.}
  \label{fig:2}
  \vspace{-0.5cm}
\end{figure}

The overall architecture of our proposed SkateFormer is depicted in Fig.~\ref{fig:2}. A skeleton sequence, corresponding to a single action, is sampled to maintain a consistent frame count of $T$, resulting in $\mathbf{X} \in \mathbb{R}^{T \times V \times M \times C_{\mathsf{in}}}$, where $V$ represents the number of joints per frame, $M$ denotes the number of individuals involved in the action, and $C_{\mathsf{in}}$ is the dimensionality of data representing a single joint ($C_{\mathsf{in}}=3$ for 3D skeletons in our cases). Initially, we reshape $\mathbf{X}$ from $(T, V, M, C_{\mathsf{in}})$ to $(T, V{\cdot}M, C_{\mathsf{in}})$ to treat joints from different individuals separately. For simplicity, we redefine $V$ as $V{\cdot}M$ so that we have $\mathbf{X} \in \mathbb{R}^{T \times V \times C_{\mathsf{in}}}$ in subsequent text.  Next, the three linear layers in the SkateFormer map the low-dimensional raw skeleton data into a higher-dimensional feature space. We then perform the skeletal-temporal positional embedding by adding learnable skeletal features and fixed temporal index features to this mapped feature. The embedded features pass through $R$ SkateFormer blocks, each comprising a self-attention layer that propagates features via skeletal-temporal relations, followed by a feed-forward layer that refines the features. The final features, after $R$ SkateFormer Blocks, undergo skeletal-temporal pooling to produce the outcome $\hat{\mathbf{y}} \in \mathbb{R}^{N_{c}}$, which is trained via backpropagation through a loss function by comparing $\hat{\mathbf{y}}$ with the true label $\mathbf{y}$, where $N_{c}$ represents the number of classes.

\subsection{SkateFormer Block}
Each of the SkateFormer Blocks in Fig.~\ref{fig:2} is structured similarly as the traditional transformer blocks \cite{Swin, Maxvit}, incorporating both a self-attention layer and a feed-forward network (FFN). The self-attention component of the SkateFormer Block can be expressed as follows:

\begin{equation}
    \begin{split}
        &[\mathbf{x}_{\mathsf{gc}}, \mathbf{x}_{\mathsf{tc}}, \mathbf{x}_{\mathsf{msa}}] = \mathsf{Split}(\mathsf{Linear}(\mathsf{LN}(\mathbf{x})))\\
        &\mathbf{x}_{\mathsf{gc}} \leftarrow \mathsf{G{\text -}Conv}(\mathbf{x}_{\mathsf{gc}})\\
        &\mathbf{x}_{\mathsf{tc}} \leftarrow \mathsf{T{\text -}Conv}(\mathbf{x}_{\mathsf{tc}})\\
        &\mathbf{x}_{\mathsf{msa}} \leftarrow \mathsf{Skate{\text -}MSA}(\mathbf{x}_{\mathsf{msa}})\\
        &\mathbf{x} \leftarrow \mathbf{x} + \mathsf{Linear}(\mathsf{Concat}(\mathbf{x}_{\mathsf{gc}}, \mathbf{x}_{\mathsf{tc}}, \mathbf{x}_{\mathsf{msa}})),
    \end{split}
    \label{eq:1}
\end{equation}

\noindent where the $\mathsf{LN}$ represents the Layer Normalization layer, the $\mathsf{Linear}$ denotes a fully-connected layer, and the $\mathsf{Split}$ indicates a channel splitting function. When input $\mathbf{x}$ to the self-attention component has $C$ channels, we split $\mathbf{x}$ into $\mathbf{x}_{\mathsf{gc}}$ of $C/4$ channels, $\mathbf{x}_{\mathsf{tc}}$ of $C/4$ channels, and $\mathbf{x}_{\mathsf{msa}}$ of $C/2$ channels. To obtain inductive biases for both skeletal and temporal aspects, we utilize a one-layer GCN ($\mathsf{G{\text -}Conv}$) and a single temporal convolution layer ($\mathsf{T{\text -}Conv}$), respectively. With a total of $H$ heads, we employ a learnable matrix of shape $(H/4, V, V)$ instead of a predefined adjacency matrix to perform $\mathsf{G{\text -}Conv}$ operations, enabling us to capture diverse connectivity patterns between joints. $\mathsf{T{\text -}Conv}$ is a 1D convolution layer with a kernel size of $k$, and performs 1D ($H/4$-group) convolutions for $\mathbf{x}_{\mathsf{tc}}$ to capture temporal dynamics. The FFN of the SkateFormer Block is expressed as follows: $\mathbf{x} \leftarrow \mathbf{x} + \mathsf{Linear}(\mathsf{Act}(\mathsf{Linear}(\mathsf{LN}(\mathbf{x}))))$, where $\mathsf{Act}$ represents the activation layer. When downsampling is needed for the input to a SkateFormer Block, a 1D convolution with a stride of 2 and Batch Normalization layer are applied.

\subsection{Skate-MSA}

\begin{figure}[htbp]
  \vspace{-0.7cm}
  \centering
  \includegraphics[width=0.7\textwidth]{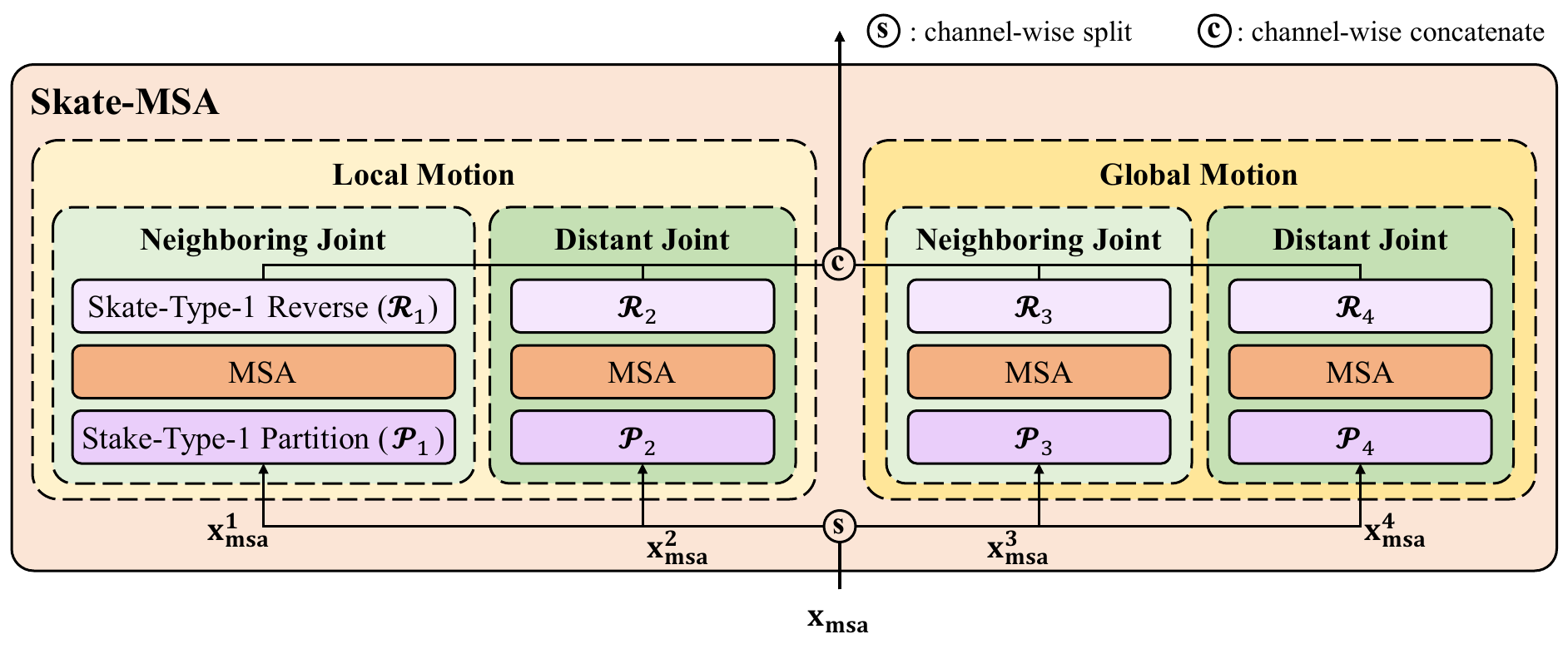}
  \vspace{-0.3cm}
  \caption{The Skate-MSA of our proposed SkateFormer.}
  \label{fig:3}
  \vspace{-0.5cm}
\end{figure}

As shown in Fig.~\ref{fig:3}, the feature map input $\mathbf{x}_{\mathsf{msa}}$ to the Skate-MSA is first split channel-wise into four equal-sized features $\mathbf{x}_{\mathsf{msa}}^{1}, \mathbf{x}_{\mathsf{msa}}^{2}, \mathbf{x}_{\mathsf{msa}}^{3}, \mathbf{x}_{\mathsf{msa}}^{4}$, each of which has $C/8$ channels. For each $\mathbf{x}_{\mathsf{msa}}^{i}$, self-attention operation is applied to discern the correlations between joints corresponding to specific relation types as:
\begin{equation}
    \begin{split}
        &\mathbf{x}_{\mathsf{msa}}^{i} \leftarrow \bm{\mathcal{R}}_{i}(\mathsf{MSA}(\bm{\mathcal{P}}_{i}(\mathbf{x}_{\mathsf{msa}}^{i}))) \\
        &\mathbf{x}_{\mathsf{msa}} \leftarrow \mathsf{Concat}(\mathbf{x}_{\mathsf{msa}}^{1}, \mathbf{x}_{\mathsf{msa}}^{2}, \mathbf{x}_{\mathsf{msa}}^{3}, \mathbf{x}_{\mathsf{msa}}^{4}),
    \end{split}
    \label{eq:2}
\end{equation}

\noindent where $\bm{\mathcal{P}}_{i}$ and $\bm{\mathcal{R}}_{i}$ represent the $i$-th Skate-Type partition and reverse operations, respectively. This approach enables the Skate-MSA to effectively analyze and model various joint relations in a specialized manner, contributing to its overall performance improvement.

\begin{wrapfigure}{r}{0.5\textwidth}
  \vspace{-0.0cm}
  \centering
  \includegraphics[width=0.48\textwidth]{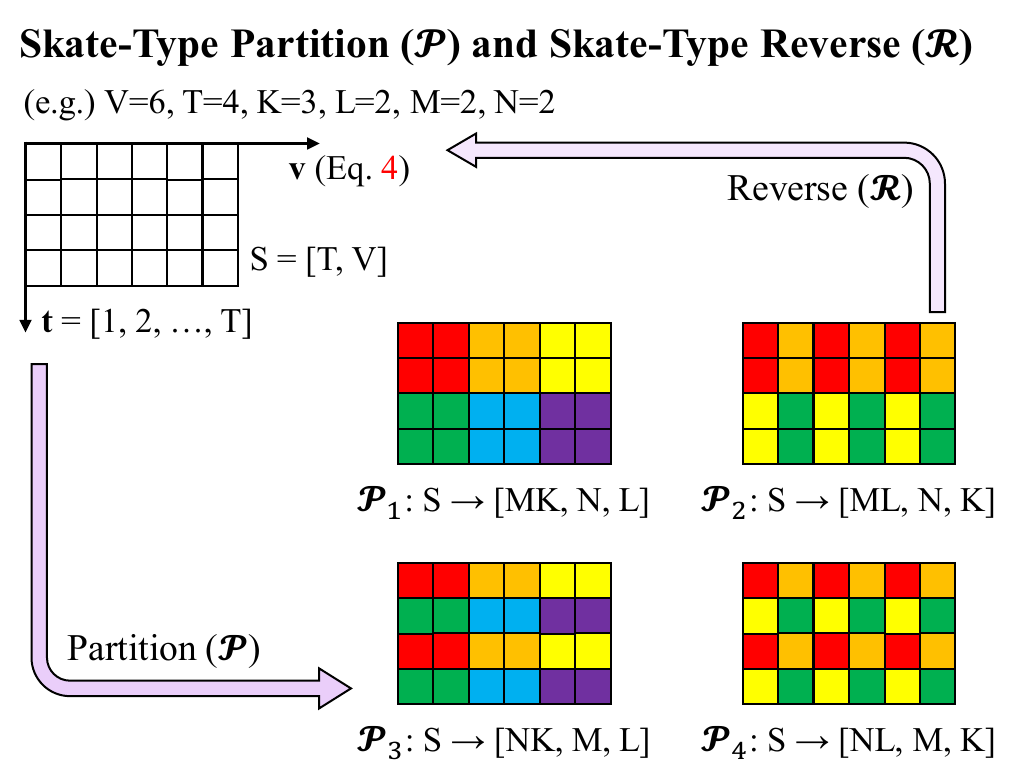}
  \caption{Skate-Type partition and reverse.}
  \label{fig:4}
  \vspace{-0.5cm}
\end{wrapfigure}

\noindent \textbf{Partition and Reverse.}\quad Let a frame be $\mathbf{f}^{t} \in \mathbb{R}^{V \times C_{\mathsf{in}}}$ at time $t$. So, we have $\mathbf{X} = \{\mathbf{f}^{t}\}_{t=1}^{T} \in \mathbb{R}^{T \times V \times C_{\mathsf{in}}}$ where $\mathbf{f}^{t}$ consists of the 3D ($C_{\mathsf{in}}=3$) positions of $V$ joints at time $t$. The frames of $\mathbf{X}$ are often temporally highly correlated for most of the action types. However, the joint positions of skeletons tend to be less correlated among some joints, depending on different action types. Nevertheless, some other joints such as the joints of the head, neck, shoulders, abdomen and pelvises are relatively not highly variant among themselves for various action types. That is, they may move together in a group. So, it is efficient to partition the whole set of joints into smaller-sized joint partitions that can be useful to distinguish different actions. Furthermore, it is also worthwhile to consider categorizing the movements of various actions into two partitions, such as local and global motions: local motion, which only changes the positions of a small number of joints with partial movements such as teeth brushing and clapping, and global motion, which appears in the entire frames and includes actions like sitting down and standing up. Motivated from this, we partition the action into four skeletal-temporal relation types. For this, we first partition the entire set of joints into total non-overlapping $K$ subsets as neighboring joint partitions $\mathbf{v}_{k}^{\mathsf{njp}}$. For example, when $K = 5$, we may have $\mathbf{v}_{1}^{\mathsf{njp}}$, $\mathbf{v}_{2}^{\mathsf{njp}}$, $\mathbf{v}_{3}^{\mathsf{njp}}$, $\mathbf{v}_{4}^{\mathsf{njp}}$ and $\mathbf{v}_{5}^{\mathsf{njp}}$ as the right arm, left arm, right leg, left leg and torso, respectively. The elements within each subset are ordered in a manner that extends outward from the body's central region (e.g., for $k = 3$, $v_{3,1}$, $v_{3,2}$ and $v_{3,3}$ are the 3D coordinates of the pelvis, right knee and right foot positions, respectively):

\begin{equation}
    \mathbf{v}_{k}^{\mathsf{njp}} = [v_{k,1}, v_{k,2}, ..., v_{k, L}],\\
    \label{eq:3}
\end{equation}

\noindent where $k=1, 2, ..., K$, $L$ represents the total number of its elements, and $\mathbf{v}_{i}^{\mathsf{njp}} \cap \mathbf{v}_{j}^{\mathsf{njp}} = \varnothing$ for $i\neq j$. We stack $\mathbf{v}_{k}^{\mathsf{njp}}$'s, thus creating the skeletal axis as:

\begin{equation}
    \mathbf{v} = \left[\mathbf{v}_{1}^{\mathsf{njp}}| \mathbf{v}_{2}^{\mathsf{njp}}| \cdots| \mathbf{v}_{K}^{\mathsf{njp}}\right],\\
    \label{eq:4}
\end{equation}

\noindent where $|\mathbf{v}| = KL = V$. Also, to consider the relations between the joints that can be physically separated in a distance each other, we group the same-positioned elements of $\mathbf{v}_{k}^{\mathsf{njp}}$ to create possibly distant joint partitions $\mathbf{v}_{l}^{\mathsf{djp}}$ as:

\begin{equation}
    \mathbf{v}_{l}^{\mathsf{djp}} = [v_{1,l}, v_{2,l}, ..., v_{K,l}],
    \label{eq:5}
\end{equation}

\noindent where $l=1, 2, ..., L$. Similarly, we define $\mathbf{t} = \left[1, 2, ..., T\right]$ as a time axis. We further define two semantic time axes as $\mathbf{t}_{m}^{\mathsf{local}}$ for local motion comprehension and $\mathbf{t}_{n}^{\mathsf{global}}$ for global motion understanding as:

\begin{equation}
    \begin{split}
        &\mathbf{t}_{m}^{\mathsf{local}} = \left[(m-1)N+1, (m-1)N+2, ..., mN\right]\\
        &\mathbf{t}_{n}^{\mathsf{global}} = \left[n, n+N, ..., n+(M-1)N\right],\\
    \end{split}
    \label{eq:6}
\end{equation}

\noindent where $m=1,2,...,M$ with the total $M$ number of $\mathbf{t}_{n}^{\mathsf{global}}$ elements, and $n=1,2,...,N$ with the total $N$ number of $\mathbf{t}_{m}^{\mathsf{local}}$ elements. Note that $\mathbf{t}_{m}^{\mathsf{local}}$ is a segment of consecutive time indices to represent a local motion within the time segment while $\mathbf{t}_{n}^{\mathsf{global}}$ is an $N$-strided sparse time axis to capture the global motion over $\mathbf{t}$. So, we have $\mathbf{t} = \{\mathbf{t}_{m}^{\mathsf{local}}\}_{m=1}^{M} = \{\mathbf{t}_{n}^{\mathsf{global}}\}_{n=1}^{N}$ and $|\mathbf{t}| = MN = T$. Based on these time axes, our skeletal-temporal partitions of joints and frames are explained in the followings.  

We partition the joints and frames together into four types in the context of skeletal-temporal relation for the Skate-MSA -- Skate-Type-1, -2, -3 and -4: (i) The Skate-Type-1 partition, denoted as $\bm{\mathcal{P}}_{1}$, pertains to a self-attention branch targeting neighboring joints and local motion, based on $\mathbf{v}_{k}^{\mathsf{njp}}$ and $\mathbf{t}_{m}^{\mathsf{local}}$; (ii) the Skate-Type-2 partition $(\bm{\mathcal{P}}_{2})$ represents a branch for distant joints and local motion, based on $\mathbf{v}_{l}^{\mathsf{djp}}$ and $\mathbf{t}_{m}^{\mathsf{local}}$; (iii) the Skate-Type-3 partition $(\bm{\mathcal{P}}_{3})$ signifies a branch for neighboring joints and global motion, based on $\mathbf{v}_{k}^{\mathsf{njp}}$ and $\mathbf{t}_{n}^{\mathsf{global}}$; (iv) Lastly, the Skate-Type-4 partition $(\bm{\mathcal{P}}_{4})$ corresponds to a branch targeting distant joints and global motion, based on $\mathbf{v}_{l}^{\mathsf{djp}}$ and $\mathbf{t}_{n}^{\mathsf{global}}$. The Skate-Type partition operations transform the shape $(S = (T, V, c))$ of $\mathbf{x}_{\mathsf{msa}}^{i}$ into: 

\begin{equation}
    \begin{split}
        \bm{\mathcal{P}}_{1}: S &\rightarrow (MK, N, L, c) \quad \bm{\mathcal{P}}_{2}: S \rightarrow (ML, N, K, c) \\
        \bm{\mathcal{P}}_{3}: S &\rightarrow (NK, M, L, c) \quad \bm{\mathcal{P}}_{4}: S \rightarrow (NL, M, K, c),
    \end{split}
    \label{eq:7}
\end{equation}

\noindent where $c = C/8$. The partitioned feature map $\mathbf{x}_{\mathsf{msa}}^{i, \bm{\mathcal{P}}} = \bm{\mathcal{P}}_{i}(\mathbf{x}_{\mathsf{msa}}^{i})$ undergoes multi-head self-attention ($\mathsf{MSA}$) and is then reshaped back to its original size of $(T, V, c)$ through a  Skate-Type reverse operation $\bm{\mathcal{R}}_{i}$ according to Eq.~\ref{eq:2}.

\noindent \textbf{Multi-head self-attention.}\quad In order to collectively consider the skeletal-temporal relation types, we can generalize feature maps $\mathbf{x}_{\mathsf{msa}}^{i, \bm{\mathcal{P}}}$ to have a shape of $(B, T^{\prime}, V^{\prime}, c)$. The feature maps are first reshaped into $(B, T^{\prime}{\cdot}V^{\prime}, c)$. Through linear mappings of $\mathbf{x}_{\mathsf{msa}}^{i, \bm{\mathcal{P}}}$ of shape $(c, c)$ by $\mathbf{W}_{Q}$, $\mathbf{W}_{K}$, and $\mathbf{W}_{V}$, we obtain the query ($\mathbf{Q}$), key ($\mathbf{K}$), and value ($\mathbf{V}$) tensors as $[\mathbf{Q}, \mathbf{K}, \mathbf{V}] = \mathbf{x}_{\mathsf{msa}}^{i, \bm{\mathcal{P}}} \cdot [\mathbf{W}_{Q}, \mathbf{W}_{K}, \mathbf{W}_{V}]$.

Since we utilized half of the total $H$ heads for $\mathsf{G{\text -}Conv}$ and $\mathsf{T{\text -}Conv}$ operations, the remaining $H/2$ heads are divided into quarters with $H^{\prime} = H/8$. They are then assigned differently with the four Skate-Types, each of which has a $\mathsf{MSA}$. We reshaped $\mathbf{Q}$, $\mathbf{K}$, $\mathbf{V}$ to have a shape of $(B, H^{\prime}, T^{\prime}{\cdot}V^{\prime}, c/H^{\prime})$. This refers to the self-attention ($\mathsf{SA}$) for each individual head $h$, given as:

\begin{equation}
    \begin{split}
        &\mathsf{SA}_{h}(\mathbf{x}_{\mathsf{msa}}^{i, \bm{\mathcal{P}}}) = \mathsf{SoftMax}(\mathbf{Q}_{h}\mathbf{K}_{h}^{\mathsf{T}}/\sqrt{c/H^{\prime}}+\mathbf{B}_{h})\mathbf{V}_{h}\\
        &\mathsf{MSA}(\mathbf{x}_{\mathsf{msa}}^{i, \bm{\mathcal{P}}}) = \mathsf{Concat}(\mathsf{SA}_{1}(\mathbf{x}_{\mathsf{msa}}^{i, \bm{\mathcal{P}}}), ..., \mathsf{SA}_{H^{\prime}}(\mathbf{x}_{\mathsf{msa}}^{i, \bm{\mathcal{P}}})),
    \end{split}
    \label{eq:8}
\end{equation}

\noindent where $\mathbf{Q}_{h}\mathbf{K}_{h}^{\mathsf{T}}$ is a data-dependent term that varies based on the input, and $\mathbf{B}_{h}$ represents the skeletal-temporal positional bias. To account for the characteristics of the temporal axis $\mathbf{t}$, we applied 1D relative positional bias: $\mathbf{B}_{h}^{\mathbf{t}} \in  \mathbb{R}^{T^{\prime} \times T^{\prime}}$ \cite{Swin} for the temporal splits, $\mathbf{t}_{m}^{\mathsf{local}}$ and $\mathbf{t}_{n}^{\mathsf{global}}$ in Eq.~\ref{eq:6}. Regarding the skeletal axis $\mathbf{v}$, for the skeletal split $\mathbf{v}_{k}^{\mathsf{njp}}$ in Eq.~\ref{eq:3}, due to the inconsistency between joints at the same position across partitions, no additional positional bias was applied: $\mathbf{B}_{h}^{\mathbf{v}}=\mathbbm{1}^{V^{\prime} \times V^{\prime}}$. For the skeletal split $\mathbf{v}_{l}^{\mathsf{djp}}$ in Eq.~\ref{eq:5}, where elements at the same position always represent a consistent semantic part of the body, we utilized absolute positional bias: $\mathbf{B}_{h}^{\mathbf{v}} \in  \mathbb{R}^{V^{\prime} \times V^{\prime}}$. The skeletal-temporal positional bias $\mathbf{B}_{h}$ is: $\mathbf{B}_{h}=\mathbf{B}_{h}^{\mathbf{t}} \otimes \mathbf{B}_{h}^{\mathbf{v}} \in  \mathbb{R}^{T^{\prime}V^{\prime} \times T^{\prime}V^{\prime}}$, where $\otimes$ is a Kronecker product.

\noindent \textbf{Complexity analysis.}\quad The computational complexity of a naive self-attention layer \cite{Attention, ViT} for feature map $\mathbf{x}_{\mathsf{msa}}$ with a shape of $(T, V, C/2)$ is $2(VT)^{2}(C/2)$. In contrast, the computation complexity of our Skate-MSA is:

\begin{equation}
    2(VT)^{2}\frac{C}{2}\left[\frac{1}{4}\left(\frac{1}{MK}+\frac{1}{ML}+\frac{1}{NK}+\frac{1}{NL}\right)\right].
    \label{eq:9}
\end{equation}
\noindent In our experimental settings, this results in approximately a 48$\times$ reduction in computational complexity compared to the naive self-attention layer.

\vspace{-0.3cm}
\subsection{Skeletal-Temporal Positional Embedding}

Our novel skeletal-temporal positional embedding method, called Skate-Embedding, is tightly associated with temporal data augmentation. So, we first explain data augmentation with our contribution and then describe the Skate-Embedding.

\noindent \textbf{Intra-instance augmentation.}\quad We define intra-instance augmentation as data augmentation within each frame sequence. To ensure stable training of transformer-based models that require a substantial amount of data, we employed various data augmentation techniques. Previous skeleton-based action recognition methods often used data augmentation by temporally sampling input frames with fixed strides or randomly for the whole input, which we refer to as \textit{temporal augmentation}. Also, as \textit{skeletal augmentation}, various transformations such as the actor order permutation, random shear, random rotation, random scaling, random coordinate drop, and random joint dropout etc. have been applied \cite{ActCLR, AimCLR}. 

\begin{figure}[tbp]
  \centering
  \includegraphics[width=0.7\textwidth]{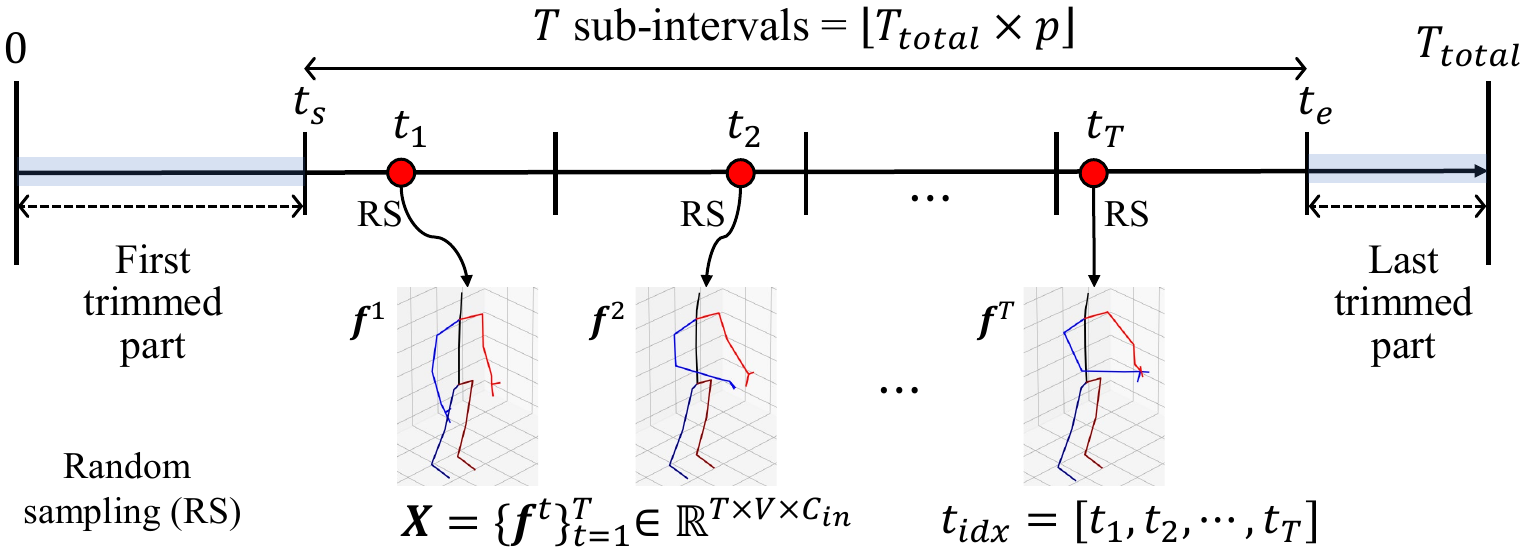}
  \caption{The proposed trimmed-uniform random sampling of frames with $p$ portion.}
  \label{fig:5}
  \vspace{-0.3cm}
\end{figure}

We propose trimmed-uniform random sampling of frames with $p$ portion. This sampling cuts out the first and last parts of the total input sequence and performs uniform random sampling of frames. Fig.~\ref{fig:5} illustrates our trimmed-uniform random sampling of frames with $p$ portion. From our trimmed-uniform random sampling of frames with a $p$ portion, the masking effect of skeleton sequences in their front and back portions, as well as a more dense sampling effect in the middle, are expected. This can lead to a concentration in the middle portion, resulting in better data augmentation.

\noindent \textbf{Inter-instance augmentation.}\quad We define inter-instance augmentation as data augmentation by exchanging the bone lengths of different subjects across different frame sequences (not within each frame sequence). By doing so, the resulting data augmentation can provide the diversity of subjects with different body sizes, which can help generalization learning.

\noindent \textbf{Skate-Embedding.}\quad We propose a novel skeletal-temporal positional embedding method, called Skate-Embedding, that utilizes fixed (not learnable) temporal index features and learnable (not fixed) skeletal features. The temporal index features are suitable for conveying to the first SkateFormer Block the temporal positional information of the sampled frames from sequences of various lengths. The sampled temporal indices are designated as $t_{\mathsf{idx}} = [t_{1}, t_{2}, ..., t_{T}]$, as illustrated in Fig~\ref{fig:5}. These temporal indices are then normalized to the range $[-1, 1]$, and are used for the fixed temporal index features as done in the temporal positional embedding. The fixed temporal index features, denoted as $\mathsf{TE} \in \mathbb{R}^{T \times C}$, are constructed for  $t_{\mathsf{idx}}$ by using the sinusoidal positional embedding \cite{Attention}. On the other hand, as skeletal joint positional (which are not the 3D coordinates of joints but their indices) embeddings, the learnable skeletal features, denoted as $\mathsf{SE} \in \mathbb{R}^{V \times C}$, are learned within the Skate-Embedding as shown in Fig.~\ref{fig:2}. Finally, the skeletal-temporal positional embedding $\mathsf{STE} \in \mathbb{R}^{T \times V \times C}$ is done by taking the outer product of $\mathsf{SE}$ and $\mathsf{TE}$ as $\mathsf{STE}[i, j, d] = \mathsf{SE}[j, d]\cdot \mathsf{TE}[i, d]$ at the $i$-th time, the $j$-th joint and the $d$-th channel.

\vspace{-0.3cm}
\section{Experimental Results}
\label{sec:exp}

\subsection{Datasets}
\noindent \textbf{NTU RGB+D.}\quad This dataset \cite{Ntu60} offers 60 action classes and includes diverse activities like drinking water, eating, brushing teeth, dropping objects, and more. It comprises total 56,880 videos captured from 40 subjects across 155 camera viewpoints. The dataset utilizes Kinect v2, encompassing RGB, IR, depth, and 3D skeleton data, and supports cross-subject (\textit{X-Sub60}) and cross-view (\textit{X-View60}) evaluation. Out of the 60 action classes, only 11 are related to human interaction, specifically when two individuals are present. We denote this subset as \textit{NTU-Inter} \cite{IGFormer, skeletr}.

\noindent \textbf{NTU RGB+D 120.}\quad This dataset \cite{Ntu120} extends the NTU RGB+D dataset \cite{Ntu60} with 120 action classes, covering actions such as putting on headphones, basketball shooting, juggling table tennis balls, and more. It contains 114,480 videos from 106 subjects across 155 camera viewpoints. It enables cross-subject (\textit{X-Sub120}) and cross-setup (\textit{X-Set120}) evaluation based on different subject and camera setups. Out of the 120 action classes, only 26 are related to human interaction. We denote this subset as \textit{NTU-Inter 120} \cite{IGFormer, skeletr, istanet}.

\noindent \textbf{NW-UCLA.}\quad This dataset \cite{NWUCLA} includes 10 action classes, such as picking up objects, dropping trash, walking, sitting, standing, and more. It comprises total 1,475 videos from 10 subjects across three camera views, collected using Kinect v1 \cite{Sensor} and featuring RGB, IR, depth, and 3D skeleton data. Evaluation follows the cross-view approach, utilizing two training views and one test view.

\vspace{-0.3cm}
\subsection{Experiment Details}

All experiments were performed using the PyTorch framework \cite{Pytorch}, running on a single NVIDIA DGX A100 GPU. Each model was trained with a total of 500 epochs. We employed a linear warm-up strategy for the learning rate, gradually increasing it from $10^{-7}$ to $10^{-3}$ during the first 25 epochs. Subsequently, a cosine-annealing scheduler \cite{CosineAnneal} was used to update the learning rate at each iteration for the remaining epochs. We employed the AdamW optimizer \cite{AdamW} with a betas of (0.9, 0.999), a weight decay of $5\times 10^{-4}$. Additionally, gradient clipping \cite{GradientClip} was applied for the loss values with gradients exceeding 1. The batch size was set to 128, the random seed was fixed at 1, and we adopted the label-smoothed cross-entropy loss \cite{LS1, LS2} as our loss function with parameter $\alpha=0.1$.

From various experiments, we found the empirical values as follows. For our experiments with the NTU RGB+D and NTU RGB+D 120 datasets, we used the following configuration: $V=48$ (excluding the center of the body, resulting in 24 joints for each of the two individuals), $T=64$, $L=4$, $K=12$, $M=8$, $N=8$, and $C=96$. For the experiments with the NW-UCLA dataset, $V=20$, $T=64$, $L=4$, $K=5$, $M=8$, $N=8$, and $C=96$.

\vspace{-0.4cm}
\subsection{Performance Comparison}

We present a comprehensive performance comparison for our SkateFormer against recent state-of-the-art skeleton-based action recognition methods. The performance comparison is made for three ensembles of different modalities ($\mathbb{E}_{1}$, $\mathbb{E}_{2}$, $\mathbb{E}_{4}$): (i) $\mathbb{E}_{1}$ - \textit{joint} modality only; (ii) $\mathbb{E}_{2}$ - \textit{joint} + \textit{bone} modalities; and (iii) $\mathbb{E}_{4}$ - \textit{joint} + \textit{bone} + \textit{joint motion} + \textit{bone motion} modalities. As in \cite{InfoGCN, FRHead, CTRGCN, Hypergraph}, we train separate networks for each modality and ensemble their outputs. Table~\ref{tab:ntu} shows the overall performance comparisons of various methods for the NTU RGB+D, NTU RGB+D 120 and NW-UCLA datasets under the three ensembles. Notably, several works address human interaction recognition \cite{IGFormer, istanet, skeletr}, a sub-part of skeleton-based action recognition, specifically focusing on scenarios where two or more individuals coexist within a single action. Accordingly, we additionally present the performance of human interaction recognition methods on the NTU Inter and NTU Inter 120 datasets in Table~\ref{tab:ntuinter}.

\begin{table}[tbp]
\scriptsize
    \centering
    \caption{Top-1 accuracy of different skeleton-based action recognition methods on the NTU RGB+D, NTU RGB+D 120 and NW-UCLA datasets. The methods of RNN, CNN, GCN, and Transformer types are evaluated based on the number of input frames and the use of ensemble strategies. The \textbf{best performances} are highlighted in \textbf{bold}.}
    \vspace{-0.1cm}
    \resizebox{\textwidth}{!}{
    \def\arraystretch{1.2}
    \begin{tabular}{c|l|c|c|c|c|c|c|c|c|c|c|c|c|c|c}
        \Xhline{2\arrayrulewidth}
        \multirow{3}{*}{Types} & \multirow{3}{*}{Methods} & \multirow{3}{*}{Frames} & \multicolumn{6}{c|}{NTU RGB+D (\%)} & \multicolumn{6}{c|}{NTU RGB+D 120 (\%)} & \multirow{3}{*}{\makecell{NW-UCLA\\(\%)}}\\
        \cline{4-15}
        & & & \multicolumn{3}{c|}{X-Sub60} & \multicolumn{3}{c|}{X-View60} & \multicolumn{3}{c|}{X-Sub120} & \multicolumn{3}{c|}{X-Set120} & \\
        \cline{4-15}
        & & & $\mathbb{E}_{1}$ & $\mathbb{E}_{2}$ & $\mathbb{E}_{4}$ & $\mathbb{E}_{1}$ & $\mathbb{E}_{2}$ & $\mathbb{E}_{4}$ & $\mathbb{E}_{1}$ & $\mathbb{E}_{2}$ & $\mathbb{E}_{4}$ & $\mathbb{E}_{1}$ & $\mathbb{E}_{2}$ & $\mathbb{E}_{4}$ & \\
        \hline
        \multirow{1}{*}{RNN} & AGC-LSTM \cite{AGCLSTM} & 100 & 87.5 & 89.2 & - & 93.5 & 95.0 & - & - & - & - & - & - & - & 93.3 \\ 
        \hline
        \multirow{2}{*}{CNN} & TA-CNN \cite{TACNN} & 64 & 88.8 & - & 90.4 & 93.6 & - & 94.8 & 82.4 & - & 85.4 & 84.0 & - & 86.8 & 96.1 \\
        & Ske2Grid \cite{Ske2Grid} & 100 & 88.3 & - & - & 95.7 & - & - & 82.7 & - & - & 85.1 & - & - & - \\
        \hline
        \multirow{9}{*}{GCN} & SGN \cite{SGN} & 20 & - & 89.0 & - & - & 94.5 & - & - & 79.2 & - & - & 81.5 & - & - \\
        & CTR-GCN \cite{CTRGCN} & 64 & 89.9 & - & 92.4 & - & - & 96.8 & 84.9 & 88.7 & 88.9 & - & 90.1 & 90.6 & 96.5 \\
        & ST-GCN++ \cite{STGCN++} & 100 & 89.3 & 91.4 & 92.1 & 95.6 & 96.7 & 97.0 & 83.2 & 87.0 & 87.5 & 85.6 & 87.5 & 89.8 & - \\
        & InfoGCN \cite{InfoGCN} & 64 & - & - & 92.7 & - & - & 96.9 & 85.1 & 88.5 & 89.4 & 86.3 & 89.7 & 90.7 & 96.6 \\
        & FR-Head \cite{FRHead} & 64 & 90.3 & 92.3 & 92.8 & 95.3 & 96.4 & 96.8 & 85.5 & - & 89.5 & 87.3 & - & 90.9 & 96.8 \\
        & Koopman \cite{koopman} & 64 & 90.2 & - & 92.9 & 95.2 & - & 96.8 & 85.7 & - & \textbf{90.0} & 87.4 & - & 91.3 & 97.0 \\
        & LST \cite{LST} & 64 & 90.2 & - & 92.9 & 95.6 & - & 97.0 & 85.5 & - & 89.9 & 87.0 & - & 91.1 & 97.2 \\
        & HD-GCN \cite{HDGCN} & 64 & 90.6 & 92.4 & 93.0 & 95.7 & 96.6 & 97.0 & 85.7 & 89.1 & 89.8 & 87.3 & 90.6 & 91.2 & 96.9 \\
        & STC-Net \cite{stcnet} & 64 & - & 92.5 & 93.0 & - & 96.7 & 97.1 & - & 89.3 & 89.9 & - & 90.7 & 91.3 & 97.2 \\
        \hline
        \multirow{5}{*}{\makecell{Trans-\\former}} & DSTA-Net \cite{DSTA} & 128 & - & - & 91.5 & - & - & 96.4 & - & - & 86.6 & - & - & 89.0 & - \\
        & STST \cite{STST} & 128 & - & - & 91.9 & - & - & 96.8 & - & - & - & - & - & - & - \\
        & FG-STFormer \cite{fgstformer} & 128 & - & - & 92.6 & - & - & 96.7 & - & - & 89.0 & - & - & 90.6 & 97.0 \\
        & Hyperformer \cite{Hypergraph} & 64 & 90.7 & - & 92.9 & 95.1 & - & 96.5 & 86.6 & - & 89.9 & 88.0 & - & 91.3 & 96.7 \\
        &\cellcolor{c_lowbest}\textbf{SkateFormer} &\cellcolor{c_lowbest}64 &\cellcolor{c_lowbest}\textbf{92.6} &\cellcolor{c_lowbest}\textbf{93.0} &\cellcolor{c_lowbest}\textbf{93.5} &\cellcolor{c_lowbest}\textbf{97.0} &\cellcolor{c_lowbest}\textbf{97.4} &\cellcolor{c_lowbest}\textbf{97.8} &\cellcolor{c_lowbest}\textbf{87.7} &\cellcolor{c_lowbest}\textbf{89.4} &\cellcolor{c_lowbest}89.8 &\cellcolor{c_lowbest}\textbf{89.3} &\cellcolor{c_lowbest}\textbf{91.0} &\cellcolor{c_lowbest}\textbf{91.4} &\cellcolor{c_lowbest}\textbf{98.3} \\
        \Xhline{2\arrayrulewidth}
    \end{tabular}}
    \label{tab:ntu}
\end{table}

\noindent \textbf{$\mathbb{E}_{1}$, $\mathbb{E}_{2}$, $\mathbb{E}_{4}$  performance.}\quad As shown in Tables~\ref{tab:ntu}, ~\ref{tab:ntuinter}, the extensive experiments demonstrate that our SkateFormer outperforms all the state-of-the-art (SoTA) methods except the $\mathbb{E}_{4}$ of X-Sub120. It should be noted that the performance of our SkateFormer is relatively higher for $\mathbb{E}_{1}$ and $\mathbb{E}_{2}$ than $\mathbb{E}_{4}$. This is indicative of the model's efficient attention strategy, which is based on partition-specific processing, allowing for the simultaneous handling of joint (input), bone (skeletal relations), joint motion (temporal relations), and bone motion (skeletal-temporal relations) inputs within a singular network framework. As the ensemble modality increases, the potential for information redundancy also rises, which may diminish the ensemble synergy. Therefore, our SkateFormer is already somewhat a strong learner in the form of a single modality compared to others.

Ensemble methods improve performance, but their effectiveness depends on the computational complexity, scaling proportionally with the number of models in the ensemble. Also, optimizing ensemble coefficients is dataset and model-specific \cite{HDGCN}, posing challenges in real-world applications. Consequently, it is crucial to leverage diverse modalities inherently within a single model to achieve both generalization and efficiency. Therefore, highlighting the importance of $\mathbb{E}_{1}$ performance is essential as a key indicator of the model's ability to generalize across diverse inputs.

\begin{table}[tbp]
    \scriptsize
    \centering
    \caption{Comparison with human interaction recognition methods on the NTU-Inter and NTU-Inter 120 datasets. The \textbf{best performances} are highlighted in \textbf{bold}.}
    \vspace{-0.1cm}
    \setlength\tabcolsep{7pt}
    \resizebox{0.99\textwidth}{!}{
    \def\arraystretch{1.2}
    \begin{tabular} {c|l|c|c|c|c|c|c|c}
        \Xhline{2\arrayrulewidth}
        \multirow{2}{*}{Types} & \multirow{2}{*}{Methods} &\multicolumn{2}{c|}{NTU-Inter ($\mathbb{E}_{1}$, \%)} & \multicolumn{2}{c|}{NTU-Inter 120 ($\mathbb{E}_{1}$, \%)} & \multirow{2}{*}{\makecell{Params.\\(M)}} & \multirow{2}{*}{\makecell{FLOPs\\(G)}} & \multirow{2}{*}{\makecell{Time\\(ms)}} \\
        \cline{3-6}
        & & X-Sub60 & X-View60 & X-Sub120 & X-Set120 & & & \\
        \hline
        \multirow{4}{*}{Transformer} & IGFormer \cite{IGFormer} & 93.6 & 96.5 & 85.4 & 86.5 & - & - & - \\
        & SkeleTR \cite{skeletr} & 94.9 & 97.7 & 87.8 & 88.3 & 3.82 & 7.30 & - \\
        & ISTA-Net \cite{istanet} & - & - & 90.6 & 91.7 & 6.22 & 68.18 & 21.71 \\
        &\cellcolor{c_lowbest}\textbf{SkateFormer} &\cellcolor{c_lowbest}\textbf{97.1} &\cellcolor{c_lowbest}\textbf{99.3} &\cellcolor{c_lowbest}\textbf{92.3} &\cellcolor{c_lowbest}\textbf{93.2} &\cellcolor{c_lowbest}\textbf{2.02} &\cellcolor{c_lowbest}\textbf{3.62} &\cellcolor{c_lowbest}\textbf{11.25} \\
        \Xhline{2\arrayrulewidth}
    \end{tabular}}
  \label{tab:ntuinter}
  \vspace{-0.7cm}
\end{table}

\begin{table}[tbp]
\scriptsize
    \centering
    \caption{Comparative analysis of SkateFormer with other methods by parameters, FLOPs, inference time, and average top-1 accuracy for joint modality.}
    \setlength\tabcolsep{7pt}
    \resizebox{0.99\textwidth}{!}{
    \def\arraystretch{1.2}
    \begin{tabular}{c|l|c|c|c|c|c}
        \Xhline{2\arrayrulewidth}
        Types & Methods & \makecell{Params.\\(M)} & \makecell{FLOPs\\(G)} & \makecell{Time\\(ms)} & \makecell{NTU RGB+D\\($\mathbb{E}_{1}$, \%)} & \makecell{NTU RGB+D 120\\($\mathbb{E}_{1}$, \%)}  \\
        \hline
        \multirow{5}{*}{GCN} & InfoGCN \cite{InfoGCN} & 1.56 & \textbf{3.34} & 12.97 & - & 85.7 \\
        & FR-Head \cite{FRHead} & \textbf{1.45} & 3.60 & 18.49 & 92.8 & 86.4 \\
        & Koopman \cite{koopman} & 5.38 & 8.76 & 17.86 & 92.7 & 86.6 \\
        & LST \cite{LST} & 2.10 & 3.60 & 18.85 & 92.9 & 86.3 \\
        & HD-GCN \cite{HDGCN} & 1.66 & 3.44 & 72.81 & 93.2 & 86.5 \\
        \hline
        \multirow{3}{*}{Transformer} & DSTA-Net \cite{DSTA} & 3.45 & 16.18 & 13.80 & - & - \\
        & Hyperformer \cite{Hypergraph} & 2.71 & 9.64 & 18.07 & 92.9 & 87.3 \\
        & \cellcolor{c_lowbest}\textbf{SkateFormer} & \cellcolor{c_lowbest}2.03 & \cellcolor{c_lowbest}3.62 & \cellcolor{c_lowbest}\textbf{11.46} & \cellcolor{c_lowbest}\textbf{94.8} & \cellcolor{c_lowbest}\textbf{88.5} \\
        \Xhline{2\arrayrulewidth}
   \end{tabular}}
   \label{tab:com}
   \vspace{-0.5cm}
\end{table}

\noindent \textbf{Computational complexity analysis.}\quad Table~\ref{tab:com} shows the complexity comparisons \cite{ptflops} for various skeleton-based action recognition methods. Our SkateFormer exhibits a competitive balance between model complexity and computational efficiency, compared to other methods. It maintains a comparable number of parameters and FLOPs with the GCN-based methods, while substantially reducing these metrics in comparison to the transformer-based counterparts. Table~\ref{tab:com} has been compiled using publicly available official code.

\vspace{-0.3cm}
\subsection{Ablation Studies}
To see the efficacy of the key components in our SkateFormer, the ablation experiments were carried for the NTU RGB+D dataset. A detailed analysis on the effectiveness of the components is provided in Tables~\ref{tab:msa}, \ref{tab:emb}, \ref{tab:aug} to quantify the impact of different design choices on the model's performance in terms of accuracy (\%) across two benchmarks: X-Sub60 and X-View60.

\noindent \textbf{Skate-Types of Skate-MSA.}\quad Table~\ref{tab:msa} presents the impact of different Skate-Types on accuracy. As shown, incorporating the skeletal relation types only improves the action classification performance over the baseline, and so does the temporal relation types only. The full model that utilizes the Skate-Types (skeletal-temporal relation types) achieves the highest accuracy, indicating that both skeletal and temporal splits are crucial in distinguishing complex actions.

\vspace{-0.5cm}
\begin{table}[h]
\scriptsize
    \centering
    \caption{Ablation study on the influence of skeletal and temporal relation types (SkateTypes) of Skate-MSA in our SkateFormer. (*: The batch size is set to 8 due to limited GPU memory.)}
    \setlength\tabcolsep{7pt}
    \resizebox{0.99\textwidth}{!}{
    \def\arraystretch{1.2}
    \begin{tabular} {l|c|c|c|c|c|c}
        \Xhline{2\arrayrulewidth}
        \multirow{2}{*}{Attention Types} & \multicolumn{2}{c|}{NTU RGB+D (\%)} & \multirow{2}{*}{\makecell{Params.\\(M)}} & \multirow{2}{*}{\makecell{FLOPs\\(G)}} & \multirow{2}{*}{\makecell{Time\\(ms)}} & \multirow{2}{*}{\makecell{Memory\\(MB)}} \\
        \cline{2-3}
        & X-Sub60 & X-View60 & & & \\
        \hline
        Baseline (no attention) & 90.7 & 95.7 & 2.03 & 3.59 & 10.68 & 80.6 \\
        + Naive self-attention* & 90.6 (\textcolor{blue}{{$\downarrow$}0.1}) & 95.0 (\textcolor{blue}{{$\downarrow$}0.7}) & 2.03 & 4.39 & 17.55 & 1612.1 \\
        \hline
        + $\mathbf{v}_{k}^{\mathsf{njp}}$ + $\mathbf{v}_{l}^{\mathsf{djp}}$ & 91.8 (\textcolor{red}{$\uparrow$1.1}) & 96.4 (\textcolor{red}{$\uparrow$0.7}) & 2.03 & 3.59 & 11.10 & 86.9 \\
        + $\mathbf{t}_{m}^{\mathsf{local}}$ + $\mathbf{t}_{n}^{\mathsf{global}}$ & 91.9 (\textcolor{red}{$\uparrow$1.2}) & 96.6 (\textcolor{red}{$\uparrow$0.9}) & 2.03 & 3.59 & 11.09 & 85.5 \\
        \hline
        + $\mathbf{v}_{k}^{\mathsf{njp}}$ + $\mathbf{v}_{l}^{\mathsf{djp}}$ + $\mathbf{t}_{m}^{\mathsf{local}}$ + $\mathbf{t}_{n}^{\mathsf{global}}$ & \textbf{92.6} (\textbf{\textcolor{red}{$\uparrow$1.9}}) & \textbf{97.0} (\textbf{\textcolor{red}{$\uparrow$1.3}}) & 2.03 & 3.62 & 11.46 & 137.6 \\
        \Xhline{2\arrayrulewidth}
    \end{tabular}}
    \label{tab:msa}
    \vspace{-0.3cm}
\end{table}

\noindent \textbf{Exploration of Skate-Embedding.}\quad Table~\ref{tab:emb} shows the ablation study on the $\mathsf{STE}$ to assess the impact of different skeletal and temporal embedding methods. We explore various combinations of skeletal and temporal embedding methods. The learnable skeletal embedding ($\mathsf{SE}$) paired with fixed temporal embedding ($\mathsf{TE}$) achieves superior performance, suggesting an optimal balance between adaptability and stability in embeddings. The Skate-Embedding allows for a more tailored feature representation, leading to improved recognition accuracy.

\begin{table}[h]
\parbox{0.445\textwidth}{
    \scriptsize
    \centering
    \caption{Ablation study results on the Skate-Embedding to assess skeletal and temporal embedding impacts.}
    \vspace{-0.3cm}
    \resizebox{0.445\textwidth}{!}{
    \def\arraystretch{1.2}
    \begin{tabular} {c|c|c|c}
        \Xhline{2\arrayrulewidth}
        \multicolumn{2}{c|}{Embedding Methods} & \multicolumn{2}{c}{NTU RGB+D (\%)}\\
        \hline
        Skeletal & Temporal & X-Sub60 & X-View60 \\
        \hline
        \multirow{3}{*}{\xmark} & \xmark & 91.9 (\textcolor{blue}{$\downarrow$0.7}) & 96.6 (\textcolor{blue}{$\downarrow$0.4}) \\
        & Learnable & 91.8 (\textcolor{blue}{$\downarrow$0.8}) & 96.5 (\textcolor{blue}{$\downarrow$0.5}) \\
        & Fixed ($\mathsf{TE}$) & 91.6 (\textcolor{blue}{$\downarrow$1.0}) & 96.6 (\textcolor{blue}{$\downarrow$0.4}) \\
        \hline
        \multirow{3}{*}{Learnable ($\mathsf{SE}$)} & \xmark & 92.1 (\textcolor{blue}{$\downarrow$0.5}) & 96.5 (\textcolor{blue}{$\downarrow$0.5}) \\
        & Learnable & 91.4 (\textcolor{blue}{$\downarrow$1.2}) & 96.2 (\textcolor{blue}{$\downarrow$0.8}) \\
        & Fixed ($\mathsf{TE}$) & \textbf{92.6} & \textbf{97.0} \\
        \Xhline{2\arrayrulewidth}
    \end{tabular}}
    \label{tab:emb}}
\;\;
\parbox{0.515\textwidth}{
    \scriptsize
    \centering
    \caption{Ablation study results on the effect of intra-instance and inter-instance augmentations.}
    \vspace{-0.3cm}
    \resizebox{0.515\textwidth}{!}{
    \def\arraystretch{1.2}
    \begin{tabular} {c|c|c|c|c}
        \Xhline{2\arrayrulewidth}
        \multicolumn{2}{c|}{Intra-instance} & \multirow{2}{*}{Inter-instance} & \multicolumn{2}{c}{NTU RGB+D (\%)}\\
        \cline{1-2} \cline{4-5}
        Temporal & Skeletal & & X-Sub60 & X-View60 \\
        \hline
        Trimmed & & & 89.8 & 94.6 \\
        Trimmed & & \cmark & 90.3 (\textcolor{red}{$\uparrow$0.5}) & 94.3 (\textcolor{blue}{$\downarrow$0.3}) \\
        Trimmed & \cmark & & 92.2 (\textcolor{red}{$\uparrow$2.4}) & 96.6 (\textcolor{red}{$\uparrow$2.0}) \\
        Fixed & \cmark & \cmark & 91.4 (\textcolor{red}{$\uparrow$1.6}) & 96.4 (\textcolor{red}{$\uparrow$1.8}) \\
        Uniform & \cmark & \cmark & 91.6 (\textcolor{red}{$\uparrow$1.8}) & 96.8 (\textcolor{red}{$\uparrow$2.2}) \\
        \hline
        Trimmed & \cmark & \cmark & \textbf{92.6} (\textbf{\textcolor{red}{$\uparrow$2.8}}) & \textbf{97.0} (\textbf{\textcolor{red}{$\uparrow$2.4}}) \\
        \Xhline{2\arrayrulewidth}
    \end{tabular}}
    \label{tab:aug}}
    \vspace{-0.3cm}
\end{table}

\noindent \textbf{Evaluation of frame sampling methods.}\quad In Table~\ref{tab:aug}, we compare three frame sampling strategies with (i) fix-strided (\textit{Fixed}), (ii) uniform random (\textit{Uniform}) and (iii) our trimmed-uniform random (\textit{Trimmed}) sampling methods. Due to the masking effect of skeleton sequences in their front and the back portions as well and more dense sampling effect in the middles during training, the trimmed-uniform random sampling is more effective in perspectives of generalization learning and data augmentation. It surpasses traditional fix-strided and uniform random sampling approaches with 1.2 (0.6)\% and 1.0 (0.2)\% margins for the X-Sub60 (X-View60), respectively.

\noindent \textbf{Impact of data augmentations.}\quad Table~\ref{tab:aug} presents the effectiveness of intra-instance (traditional) and inter-instance (additional) data augmentations. The intra-instance augmentation alone improves the performance while performance drop is observed with the inter-instance augmentation alone for the X-View60. This performance drop with 0.3\%-point in accuracy is due to the inherent characteristics of the cross-view setting such that the variability in bone lengths across different subjects within the X-View60 is limited. This implies that the model may already encapsulate a comprehensive representation of these attributes. Consequently, the introduction of inter-instance data augmentation does not contribute additional discriminative information and may instead introduce redundancy, leading to a performance decline. However, the amalgamation of both intra- and inter-instance augmentations provides a synergistic enhancement, especially by random scaling and shearing on bone lengths in the skeletal augmentation, achieving peak accuracy of 92.6\% on X-Sub60 and 97.0\% on X-View60.

\vspace{-0.3cm}
\section{Conclusions}
\label{sec:con}
\vspace{-0.1cm}
In this paper, we presented SkateFormer -- a novel skeletal-temporal transformer tailored for action recognition tasks. For this, we propose an effective partition-specific attention strategy in skeleton-based action recognition, both to capture essential features and to reduce computational complexity. For efficient training, our novel Skate-Embedding that combines skeletal and temporal features is presented, significantly enhancing action recognition performance by forming an outer product between learnable skeletal features and fixed temporal index features. Our SkateFormer sets a new state-of-the-art for action recognition performance across multiple modalities (4-ensemble condition) and single modalities (joints, bones, joint motions, bone motions), showing notable improvement over the most recent state-of-the-art methods.

\vspace{0.5cm}

\noindent \textbf{Acknowledgements.}\quad This work was supported by IITP grant funded by the Korea government(MSIT) (No.RS2022-00144444, Deep Learning Based Visual Representational Learning and Rendering of Static and Dynamic Scenes).

\bibliographystyle{splncs04}
\bibliography{main}

\clearpage
\appendix

\setcounter{figure}{5}
\setcounter{table}{6}
\setcounter{equation}{9}

\section*{\LARGE Supplementary Material}
\label{supp}

\vspace{-0.8cm}
\begin{table}[htbp]
\scriptsize
\setlength\tabcolsep{7pt}
    \centering
    \resizebox{\columnwidth}{!}{
    \def\arraystretch{1.5}
    \begin{tabular} {l|l}
        \Xhline{2\arrayrulewidth}
        \cellcolor{c_highbest}\textit{Supple.} Sec.  & \cellcolor{c_highbest}Descriptions for analysis and discussion \\
        \hline
        Sec.~\ref{sec:novelty} & \textbf{High-level novelty compared to conventional methods}\\
        \hline
        \multirow{2}{*}{Sec.~\ref{sec:b1}} & \textbf{Statistical analysis of Skate-MSA for different classes} \\
        & \makecell[l]{(i) Accuracy comparisons of SkateFormer for different classes (Table.~\ref{tab:class})\\
                (ii) Statistical analysis of action classes where Skate-MSA enhances performance \\
                (iii) Discussion of failure action classes} \\
        \hline
        Sec.~\ref{sec:b2} & \textbf{Generalization issues to diverse modality inputs} \\
        \hline
        \multirow{2}{*}{Sec.~\ref{sec:b3}} & \textbf{Visualizing the activation levels of Skate-MSA} \\
        & \makecell[l]{(i) Visualization of activation for different classes (Fig.~\ref{fig:score})\\
        (ii) Discussion of adaptive nature of Skate-MSA concerning action classes}\\
        \hline
        Sec.~\ref{sec:e1} & \textbf{Theoretical calculation of the computational complexity of Skate-MSA} \\
        \hline
        Sec.~\ref{sec:e2} & \textbf{Additional experiments on the computational complexity} \\
        \Xhline{2\arrayrulewidth}
    \end{tabular}
    }
   \label{tab:supple}
   \vspace{-0.6cm}
\end{table}

Due to the space limit of the main paper, we had a difficulty to contain all the detailed descriptions for the analysis and discussion on the various experimental results in the same granularity levels of details. Instead, high level descriptions are included in the main paper while detailed descriptions are in the \textit{Supple}. as shown in the above table.

\vspace{-0.3cm}

\section{Discussion on Various Partition-Based Approaches}
\label{sec:novelty}

In recent studies, several partition-based methods such as DSTA-Net \cite{DSTA}, ST-TR \cite{STTR}, IIP-Trnasformer \cite{iip}, STST \cite{STST}, FG-STFormer \cite{fgstformer}, Hyperformer \cite{Hypergraph}, IGFormer \cite{IGFormer}, SkeleTR \cite{skeletr}, and ISTA-Net \cite{istanet} have been proposed. As shown in Table~\ref{tab:novelty}, we introduce SkateFormer, which differs fundamentally in the following aspects:

\vspace{-0.4cm}
\begin{table}[htbp]
\scriptsize
    \centering
    \caption{Comparison with ours SkateFormer with existing transformer-based methods. S-Attn, T-Attn, T-Conv, and ST-Attn indicate `skeletal attention', `temporal attention', `temporal convolution' and `skeletal-temporal attention', respectively.}
    \vspace{-0.2cm}
    \setlength\tabcolsep{7pt}
    \resizebox{1.0\textwidth}{!}{
    \def\arraystretch{1.2}
    \begin{tabular}{c|l|c|c|c}
        \Xhline{2\arrayrulewidth}
        Tasks & Methods & Partition Types & \makecell{Tokenization\\for Attention} & Attention Types \\
        \hline
        \multirow{6}{*}{\makecell{Human\\Action\\Recognition}} & DSTA-Net \cite{DSTA} & N/A (Reshaping) & No & S-Attn, T-Attn (Sequential) \\
        & ST-TR \cite{STTR} & N/A (Reshaping)  & No & S-Attn, T-Attn (Parallel) \\
        & IIP-Transformer \cite{iip} & S-Type & Yes & S-Attn, T-Attn (Sequential) \\
        & STST \cite{STST} & N/A (Reshaping) & No & S-Attn, T-Attn (Parallel) \\
        & FG-STFormer \cite{fgstformer} & S-Type & Yes & S-Attn, T-Attn (Sequential) \\
        & Hyperformer \cite{Hypergraph} & S-Type & Yes & S-Attn, T-Conv (Sequential) \\
        \hline
        \multirow{3}{*}{\makecell{Human\\Interaction\\Recognition}} & IGFormer \cite{IGFormer} & S-Type, T-Type & Yes & Joint-group-level ST-Attn \\
        & SkeleTR \cite{skeletr} & N/A (Pooling) & Yes & Joint-group-level ST-Attn \\
        & ISTA-Net \cite{istanet} & S-Type, T-Type & Yes & Joint-group-level ST-Attn \\
        \hline
        \cellcolor{c_lowbest}Both &\cellcolor{c_lowbest}\textbf{SkateFormer} &\cellcolor{c_lowbest}4 Skate-Types &\cellcolor{c_lowbest}No &\cellcolor{c_lowbest} Joint-element-level ST-Attn \\
        \Xhline{2\arrayrulewidth}
   \end{tabular}}
   \label{tab:novelty}
\end{table}

\vspace{-0.5cm}
\begin{itemize}
\item Unlike existing methods that rely solely on physically neighboring joints (S-Type) or local motion (T-Type), our SkateFormer leverages four skeletal-temporal relation types (Skate-Types) with joint partitions (physically `neighboring' + `distant') and frame partitions (`local' + `global' motions).
\item While prior approaches typically tokenize joints within the same partition into a single feature (partition-based tokenization, not partition-specific attention), our SkateFormer adopts self-attention within each partition, enhancing its capability for fine-grained analysis.
\item To mitigate computational complexity, existing methods often employ strategies such as conducting attention solely in the skeletal dimension while utilizing temporal convolution in the temporal dimension, or separately performing skeletal and temporal attention mechanisms, or employing joint-group-level attention with tokenization. In contrast, our SkateFormer is specifically designed to efficiently capture skeletal-temporal relations at the joint-element-level, thus eliminating the need for tokenization.
\end{itemize}

\section{Additional Results}

\subsection{Recognition Accuracy based on Action Labels}
\label{sec:b1}

Table~\ref{tab:class} presents the top-1 accuracy results for single-joint modality (\textit{J}) action recognition in the NTU RGB+D X-Sub60 evaluation, categorized by action labels. The baseline model represents a model without the application of partition-specific attention based on four skeletal-temporal relation types from Skate-MSA (no attention, only $\mathbf{V}_{h}$ in Eq.~\ref{eq:8}). The model denoted as (+ $\mathbf{v}_{k}^{\mathsf{njp}}$ + $\mathbf{v}_{l}^{\mathsf{djp}}$) applies partition-specific attention based on two skeletal relation types, while (+ $\mathbf{t}_{m}^{\mathsf{local}}$ + $\mathbf{t}_{n}^{\mathsf{global}}$) utilizes partition-specific attention based on two temporal relation types. The model labeled (+$\mathbf{v}_{k}^{\mathsf{njp}}$ + $\mathbf{v}_{l}^{\mathsf{djp}}$ + $\mathbf{t}_{m}^{\mathsf{local}}$ + $\mathbf{t}_{n}^{\mathsf{global}}$) represents the full SkateFormer model, which incorporates partition-specific attention based on four Skate-Types.

Our baseline model already outperforms state-of-the-art methods \cite{STGCN, CTRGCN, STGCN++, InfoGCN, FRHead, LST, HDGCN}, as evident from the average top-1 accuracy. The \textit{Rank} in the baseline indicates the ranking of top-1 accuracy among 60 classes. Comparing the full model to the baseline, we observed performance improvements in the majority of action labels (42 out of 60 classes), with eight classes maintaining the same performance, and a slight decrease in performance for ten classes. This can be interpreted as an effective utilization of limited model capacity, where slight decreases in performance for high-performing action labels (high-rank) allow for significant improvements in performance for lower-performing action labels (low-rank).

Notably, the baseline performance for actions such as `wear a shoe' and `take off a shoe' increased substantially from (60.81\%, 78.83\%) to (87.55\%, 83.94\%), demonstrating that our partition-specific attention enhances discriminative capability for similar classes. Furthermore, performance improvements for relatively low-rank action labels, such as `typing on a keyboard', `sneeze/cough', `use a fan (with hand or paper)/feeling warm', were substantial.

\noindent \textbf{Analysis of failure cases.}\quad As shown in Table~\ref{tab:class}, most of the failure cases occur when action classes are determined by fine finger motions, e.g. `reading', `writing', `playing with phone/tablet', and `typing on a keyboard'.

\begin{table*}[tbp]
\scriptsize
    \centering
    \caption{Top-1 accuracy based on action labels in NTU RGB+D X-Sub60 evaluation.}
    \setlength\tabcolsep{7pt}
    \resizebox{\textwidth}{!}{
    \def\arraystretch{1.2}
    \begin{tabular} {l|c|c|c|c}
        \Xhline{2\arrayrulewidth}
        & & & & \\[-1em] 
        Action Label & Baseline (Rank) & +$\mathbf{v}_{k}^{\mathsf{njp}}$ + $\mathbf{v}_{l}^{\mathsf{djp}}$ & + $\mathbf{t}_{m}^{\mathsf{local}}$ + $\mathbf{t}_{n}^{\mathsf{global}}$ & +$\mathbf{v}_{k}^{\mathsf{njp}}$ + $\mathbf{v}_{l}^{\mathsf{djp}}$ + $\mathbf{t}_{m}^{\mathsf{local}}$  + $\mathbf{t}_{n}^{\mathsf{global}}$ \\
        & & & & \\[-1em] 
        \hline
        drink water & 85.77 (47) & 85.77 (\textcolor{gray}{--0.00}) & 85.77 (\textcolor{gray}{--0.00}) & 88.32 (\textcolor{red}{$\uparrow$2.55}) \\ 
        eat meal/snack & 76.00 (56) & 77.82 (\textcolor{red}{$\uparrow$1.82}) & 77.45 (\textcolor{red}{$\uparrow$1.45}) & 78.18 (\textcolor{red}{$\uparrow$2.18}) \\ 
        brushing teeth & 90.11 (42) & 85.35 (\textcolor{blue}{$\downarrow$4.76}) & 91.21 (\textcolor{red}{$\uparrow$1.10}) & 89.01 (\textcolor{blue}{$\downarrow$1.10}) \\ 
        brushing hair & 89.01 (45) & 91.21 (\textcolor{red}{$\uparrow$2.20}) & 93.77 (\textcolor{red}{$\uparrow$4.76}) & 93.41 (\textcolor{red}{$\uparrow$4.40}) \\ 
        drop & 92.00 (39) & 94.18 (\textcolor{red}{$\uparrow$2.18}) & 92.36 (\textcolor{red}{$\uparrow$0.36}) & 92.73 (\textcolor{red}{$\uparrow$0.73}) \\ 
        pickup & 97.09 (18) & 97.82 (\textcolor{red}{$\uparrow$0.73}) & 97.09 (\textcolor{gray}{--0.00}) & 98.55 (\textcolor{red}{$\uparrow$1.45}) \\ 
        throw & 92.73 (36) & 93.45 (\textcolor{red}{$\uparrow$0.73}) & 92.00 (\textcolor{blue}{$\downarrow$0.73}) & 93.82 (\textcolor{red}{$\uparrow$1.09}) \\ 
        sitting down & 98.17 (13) & 98.53 (\textcolor{red}{$\uparrow$0.37}) & 99.27 (\textcolor{red}{$\uparrow$1.10}) & 98.53 (\textcolor{red}{$\uparrow$0.37}) \\ 
        standing up (from sitting position) & 98.90 (7) & 98.53 (\textcolor{blue}{$\downarrow$0.37}) & 99.63 (\textcolor{red}{$\uparrow$0.73}) & 98.90 (\textcolor{gray}{--0.00}) \\ 
        clapping & 83.88 (51) & 81.68 (\textcolor{blue}{$\downarrow$2.20}) & 86.08 (\textcolor{red}{$\uparrow$2.20}) & 85.35 (\textcolor{red}{$\uparrow$1.47}) \\ 
        reading & 58.61 (60) & 60.81 (\textcolor{red}{$\uparrow$2.20}) & 61.54 (\textcolor{red}{$\uparrow$2.93}) & 63.37 (\textcolor{red}{$\uparrow$4.76}) \\ 
        writing & 68.38 (57) & 68.01 (\textcolor{blue}{$\downarrow$0.37}) & 66.91 (\textcolor{blue}{$\downarrow$1.47}) & 71.32 (\textcolor{red}{$\uparrow$2.94}) \\ 
        tear up paper & 95.94 (21) & 96.68 (\textcolor{red}{$\uparrow$0.74}) & 94.46 (\textcolor{blue}{$\downarrow$1.48}) & 95.94 (\textcolor{gray}{--0.00}) \\ 
        wear jacket & 98.18 (12) & 98.18 (\textcolor{gray}{--0.00}) & 98.55 (\textcolor{red}{$\uparrow$0.36}) & 97.82 (\textcolor{blue}{$\downarrow$0.36}) \\ 
        take off jacket & 98.19 (10) & 98.91 (\textcolor{red}{$\uparrow$0.72}) & 99.64 (\textcolor{red}{$\uparrow$1.45}) & 99.64 (\textcolor{red}{$\uparrow$1.45}) \\ 
        wear a shoe & 60.81 (59) & 86.81 (\textcolor{red}{$\uparrow$26.01}) & 84.62 (\textcolor{red}{$\uparrow$23.81}) & 87.55 (\textbf{\textcolor{red}{$\uparrow$26.74}}) \\ 
        take off a shoe & 78.83 (53) & 82.48 (\textcolor{red}{$\uparrow$3.65}) & 78.47 (\textcolor{blue}{$\uparrow$0.36}) & 83.94 (\textbf{\textcolor{red}{$\uparrow$5.11}}) \\ 
        wear on glasses & 93.04 (34) & 93.41 (\textcolor{red}{$\uparrow$0.37}) & 93.77 (\textcolor{red}{$\uparrow$0.73}) & 93.41 (\textcolor{red}{$\uparrow$0.37}) \\ 
        take off glasses & 95.62 (23) & 95.62 (\textcolor{gray}{--0.00}) & 95.26 (\textcolor{blue}{$\downarrow$0.36}) & 94.89 (\textcolor{blue}{$\downarrow$0.73}) \\ 
        put on a hat/cap & 96.69 (19) & 97.43 (\textcolor{red}{$\uparrow$0.74}) & 98.16 (\textcolor{red}{$\uparrow$1.47}) & 97.79 (\textcolor{red}{$\uparrow$1.10}) \\ 
        take off a hat/cap & 98.90 (7) & 98.53 (\textcolor{blue}{$\downarrow$0.37}) & 98.90 (\textcolor{gray}{--0.00}) & 98.90 (\textcolor{gray}{--0.00}) \\ 
        cheer up & 93.80 (29) & 92.34 (\textcolor{blue}{$\downarrow$1.46}) & 93.80 (\textcolor{gray}{--0.00}) & 94.89 (\textcolor{red}{$\uparrow$1.09}) \\ 
        hand waving & 93.80 (29) & 94.16 (\textcolor{red}{$\uparrow$0.36}) & 93.80 (\textcolor{gray}{--0.00}) & 93.07 (\textcolor{blue}{$\downarrow$0.73}) \\ 
        kicking something & 94.20 (27) & 94.93 (\textcolor{red}{$\uparrow$0.72}) & 97.10 (\textcolor{red}{$\uparrow$2.90}) & 97.83 (\textcolor{red}{$\uparrow$3.62}) \\ 
        reach into pocket & 84.67 (49) & 84.31 (\textcolor{blue}{$\downarrow$0.36}) & 86.13 (\textcolor{red}{$\uparrow$1.46}) & 86.86 (\textcolor{red}{$\uparrow$2.19}) \\ 
        hopping (one foot jumping) & 98.91 (6) & 98.91 (\textcolor{gray}{--0.00}) & 98.91 (\textcolor{gray}{--0.00}) & 98.91 (\textcolor{gray}{--0.00}) \\ 
        jump up & 100.00 (1) & 100.00 (\textcolor{gray}{--0.00}) & 100.00 (\textcolor{gray}{--0.00}) & 100.00 (\textcolor{gray}{--0.00}) \\ 
        make a phone call/answer phone & 90.18 (41) & 89.09 (\textcolor{blue}{$\downarrow$1.09}) & 90.55 (\textcolor{red}{$\uparrow$0.36}) & 92.00 (\textcolor{red}{$\uparrow$1.82}) \\ 
        playing with phone/tablet & 76.36 (55) & 76.36 (\textcolor{gray}{--0.00}) & 77.09 (\textcolor{red}{$\uparrow$0.73}) & 74.91 (\textcolor{blue}{$\downarrow$1.45}) \\ 
        typing on a keyboard & 64.36 (58) & 72.73 (\textcolor{red}{$\uparrow$8.36}) & 74.55 (\textcolor{red}{$\uparrow$10.18}) & 77.82 (\textbf{\textcolor{red}{$\uparrow$13.45}}) \\ 
        pointing to something with finger & 79.71 (52) & 88.04 (\textcolor{red}{$\uparrow$8.33}) & 81.88 (\textcolor{red}{$\uparrow$2.17}) & 84.42 (\textcolor{red}{$\uparrow$4.71}) \\ 
        taking a selfie & 92.39 (38) & 93.84 (\textcolor{red}{$\uparrow$1.45}) & 94.57 (\textcolor{red}{$\uparrow$2.17}) & 93.12 (\textcolor{red}{$\uparrow$0.72}) \\ 
        check time (from watch) & 93.12 (33) & 91.67 (\textcolor{blue}{$\downarrow$1.45}) & 92.75 (\textcolor{blue}{$\downarrow$0.36}) & 92.03 (\textcolor{blue}{$\downarrow$1.09}) \\ 
        rub two hands together & 89.13 (43) & 89.49 (\textcolor{red}{$\uparrow$0.36}) & 91.30 (\textcolor{red}{$\uparrow$2.17}) & 92.03 (\textcolor{red}{$\uparrow$2.90}) \\ 
        nod head/bow & 97.46 (14) & 98.55 (\textcolor{red}{$\uparrow$1.09}) & 98.55 (\textcolor{red}{$\uparrow$1.09}) & 99.28 (\textcolor{red}{$\uparrow$1.81}) \\ 
        shake head & 94.55 (26) & 95.27 (\textcolor{red}{$\uparrow$0.73}) & 96.73 (\textcolor{red}{$\uparrow$2.18}) & 96.00 (\textcolor{red}{$\uparrow$1.45}) \\ 
        wipe face & 85.87 (46) & 87.32 (\textcolor{red}{$\uparrow$1.45}) & 87.68 (\textcolor{red}{$\uparrow$1.81}) & 90.94 (\textcolor{red}{$\uparrow$5.07}) \\ 
        salute & 93.48 (31) & 93.48 (\textcolor{gray}{--0.00}) & 93.84 (\textcolor{red}{$\uparrow$0.36}) & 94.93 (\textcolor{red}{$\uparrow$1.45}) \\ 
        put the palms together & 97.46 (14) & 97.46 (\textcolor{gray}{--0.00}) & 96.74 (\textcolor{blue}{$\downarrow$0.72}) & 96.38 (\textcolor{blue}{$\downarrow$1.09}) \\ 
        cross hands in front (say stop) & 96.01 (20) & 97.10 (\textcolor{red}{$\uparrow$1.09}) & 96.38 (\textcolor{red}{$\uparrow$0.36}) & 97.46 (\textcolor{red}{$\uparrow$1.45}) \\ 
        sneeze/cough & 78.62 (54) & 82.97 (\textcolor{red}{$\uparrow$4.35}) & 83.33 (\textcolor{red}{$\uparrow$4.71}) & 85.51 (\textbf{\textcolor{red}{$\uparrow$6.88}}) \\ 
        staggering & 99.28 (4) & 99.64 (\textcolor{red}{$\uparrow$0.36}) & 100.00 (\textcolor{red}{$\uparrow$0.72}) & 99.28 (\textcolor{gray}{--0.00}) \\ 
        falling & 99.64 (2) & 100.00 (\textcolor{red}{$\uparrow$0.36}) & 99.64 (\textcolor{gray}{--0.00}) & 100.00 (\textcolor{red}{$\uparrow$0.36}) \\ 
        touch head (headache) & 84.06 (50) & 89.13 (\textcolor{red}{$\uparrow$5.07}) & 84.06 (\textcolor{gray}{--0.00}) & 88.41 (\textcolor{red}{$\uparrow$4.35}) \\ 
        touch chest (stomachache/heart pain) & 93.84 (28) & 94.57 (\textcolor{red}{$\uparrow$0.72}) & 93.48 (\textcolor{blue}{$\downarrow$0.36}) & 96.01 (\textcolor{red}{$\uparrow$2.17}) \\ 
        touch back (backache) & 95.29 (24) & 94.57 (\textcolor{blue}{$\downarrow$0.72}) & 95.29 (\textcolor{gray}{--0.00}) & 95.65 (\textcolor{red}{$\uparrow$0.36}) \\ 
        touch neck (neckache) & 90.22 (40) & 90.58 (\textcolor{red}{$\uparrow$0.36}) & 92.39 (\textcolor{red}{$\uparrow$2.17}) & 93.12 (\textcolor{red}{$\uparrow$2.90}) \\ 
        nausea or vomiting condition & 85.45 (48) & 83.64 (\textcolor{blue}{$\downarrow$1.82}) & 84.00 (\textcolor{blue}{$\downarrow$1.45}) & 85.45 (\textcolor{gray}{--0.00}) \\ 
        use a fan (with hand or paper)/feeling warm & 89.09 (44) & 90.91 (\textcolor{red}{$\uparrow$1.82}) & 89.45 (\textcolor{red}{$\uparrow$0.36}) & 94.18 (\textbf{\textcolor{red}{$\uparrow$5.09}}) \\ 
        punching/slapping other person & 92.70 (37) & 93.07 (\textcolor{red}{$\uparrow$0.36}) & 92.70 (\textcolor{gray}{--0.00}) & 94.16 (\textcolor{red}{$\uparrow$1.46}) \\ 
        kicking other person & 95.65 (22) & 96.38 (\textcolor{red}{$\uparrow$0.72}) & 95.65 (\textcolor{gray}{--0.00}) & 96.38 (\textcolor{red}{$\uparrow$0.72}) \\ 
        pushing other person & 98.55 (9) & 97.83 (\textcolor{blue}{$\downarrow$0.72}) & 97.46 (\textcolor{blue}{$\downarrow$1.09}) & 98.19 (\textcolor{blue}{$\downarrow$0.36}) \\ 
        pat on back of other person & 93.48 (31) & 95.29 (\textcolor{red}{$\uparrow$1.81}) & 94.93 (\textcolor{red}{$\uparrow$1.45}) & 95.65 (\textcolor{red}{$\uparrow$2.17}) \\ 
        point finger at the other person & 92.75 (35) & 94.20 (\textcolor{red}{$\uparrow$1.45}) & 94.57 (\textcolor{red}{$\uparrow$1.81}) & 94.20 (\textcolor{red}{$\uparrow$1.45}) \\ 
        hugging other person & 99.27 (5) & 99.27 (\textcolor{gray}{--0.00}) & 99.64 (\textcolor{red}{$\uparrow$0.36}) & 99.64 (\textcolor{red}{$\uparrow$0.36}) \\ 
        giving something to other person & 95.29 (24) & 96.74 (\textcolor{red}{$\uparrow$1.45}) & 95.65 (\textcolor{red}{$\uparrow$0.36}) & 95.65 (\textcolor{red}{$\uparrow$0.36}) \\ 
        touch other person's pocket & 97.45 (17) & 97.45 (\textcolor{gray}{--0.00}) & 98.55 (\textcolor{red}{$\uparrow$1.09}) & 96.00 (\textcolor{blue}{$\downarrow$1.45}) \\ 
        handshaking & 97.46 (14) & 97.10 (\textcolor{blue}{$\downarrow$0.36}) & 97.46 (\textcolor{gray}{--0.00}) & 97.46 (\textcolor{gray}{--0.00}) \\ 
        walking towards each other & 99.63 (3) & 100.00 (\textcolor{red}{$\uparrow$0.37}) & 100.00 (\textcolor{red}{$\uparrow$0.37}) & 100.00 (\textcolor{red}{$\uparrow$0.37}) \\ 
        walking apart from each other & 98.19 (10) & 97.10 (\textcolor{blue}{$\downarrow$1.09}) & 96.74 (\textcolor{blue}{$\downarrow$1.45}) & 97.83 (\textcolor{blue}{$\downarrow$0.36}) \\ 
        \hline
        average & 90.65 & 91.79 (\textcolor{red}{$\uparrow$1.14}) & 91.88 (\textcolor{red}{$\uparrow$1.23}) & 92.62 (\textcolor{red}{$\uparrow$1.98}) \\
        \Xhline{2\arrayrulewidth}
    \end{tabular}}
    \label{tab:class}
\end{table*}

\subsection{Single Modality Comparision}
\label{sec:b2}

In Tables~\ref{tab:sinntu60}, \ref{tab:sinntu120}, we present the results of action recognition based on single modalities. \textit{J} denotes the joint modality, \textit{B} represents the bone modality, \textit{JM} indicates joint motion modality, and \textit{BM} signifies bone motion modality. The top-1 accuracy is reported based on both the published paper and the official code provided. As an exception, for studies marked with (*), we relied on the performance reported in \cite{STGCN++}, as the official code and paper did not provide modality-specific performance. Table~\ref{tab:sinntu60} shows the top-1 accuracy for X-Sub60 and X-View60 evaluation on the NTU RGB+D \cite{Ntu60} dataset. Table~\ref{tab:sinntu120} displays the top-1 accuracy for X-Sub120 and X-Set120 evaluation on the NTU RGB+D 120 \cite{Ntu120} dataset.

In the domain of skeleton-based action recognition, modality ensemble is a common approach where networks with the same architecture are trained separately for different modalities, and the final label is determined through a weighted summation of network outcomes \cite{InfoGCN, FRHead, CTRGCN, Hypergraph}. As a consequence, the computational requirements, including FLOPs, parameter count, and inference time, increase in proportion to the number of modalities being ensembled. Therefore, achieving effective recognition performance with just a single modality is of paramount importance for real-world applications. As evident from Tables~\ref{tab:sinntu60}, \ref{tab:sinntu120}, our results showcase a significant dominance in single-modality performance over existing state-of-the-art methods, particularly in the \textit{J} and \textit{B} modalities. Furthermore, our findings demonstrate that a single modality alone can achieve performance comparable to that of ensemble modalities in previous methods.

\begin{table*}[htbp]
\scriptsize
    \centering
    \caption{Top-1 classification accuracy of different skeleton-based action recognition methods on NTU RGB+D \cite{Ntu60} dataset. The \textbf{best performances} are highlighted in \textbf{bold}.}
    \setlength\tabcolsep{7pt}
    \resizebox{\textwidth}{!}{
    \def\arraystretch{1.2}
    \begin{tabular} {l|c|c|c|c|c|c|c|c|c}
        \Xhline{2\arrayrulewidth}
        \multirow{3}{*}{Methods} & \multirow{3}{*}{Frames} & \multicolumn{8}{c}{NTU RGB+D (\%)} \\
        \cline{3-10}
        & & \multicolumn{4}{c|}{X-Sub60} & \multicolumn{4}{c}{X-View60} \\
        \cline{3-10}
        & & J & B & JM & BM & J & B & JM & BM \\
        \hline
        ST-GCN (*) \cite{STGCN} & 100 & 87.8 & 88.6 & 85.8 & 86.2 & 95.5 & 95.0 & 93.7 & 92.8 \\
        CTR-GCN \cite{CTRGCN} & 64 & 89.9 & 90.6 & 88.1 & 87.9 & - & - & - & - \\
        CTR-GCN (*) \cite{CTRGCN} & 100 & 89.6 & 90.0 & 88.0 & 87.5 & 95.6 & 95.4 & 94.4 & 93.6 \\
        ST-GCN++ \cite{STGCN++} & 100 & 89.3 & 90.1 & 87.5 & 87.3 & 95.6 & 95.5 & 94.3 & 93.8 \\
        InfoGCN \cite{InfoGCN} & 64 & 89.8 & 90.6 & 88.9 & 88.6 & 95.2 & 95.5 & 94.2 & 93.6 \\
        FR-Head \cite{FRHead} & 64 & 90.3 & 91.1 & 88.7 & 87.6 & 95.3 & 95.0 & 93.6 & 92.6 \\
        LST \cite{LST} & 64 & 90.2 & 91.2 & 88.0 & 87.8 & 95.6 & 95.5 & 93.7 & 93.2 \\
        HD-GCN \cite{HDGCN} & 64 & 90.6 & 90.9 & - & - & 95.7 & 95.1 & - & - \\
        \hline
        \cellcolor{c_lowbest}\textbf{SkateFormer} &\cellcolor{c_lowbest}64 &\cellcolor{c_lowbest}\textbf{92.6} &\cellcolor{c_lowbest}\textbf{92.1} &\cellcolor{c_lowbest}\textbf{89.8} &\cellcolor{c_lowbest}\textbf{89.0} &\cellcolor{c_lowbest}\textbf{97.0} &\cellcolor{c_lowbest}\textbf{96.5} &\cellcolor{c_lowbest}\textbf{95.8} &\cellcolor{c_lowbest}\textbf{94.7} \\
        \Xhline{2\arrayrulewidth}
    \end{tabular}}
  \label{tab:sinntu60}
\end{table*}

\begin{table*}[htbp]
\scriptsize
    \centering
    \caption{Top-1 classification accuracy of different skeleton-based action recognition methods on NTU RGB+D 120 \cite{Ntu120} dataset. The \textbf{best performances} are highlighted in \textbf{bold}.}
    \setlength\tabcolsep{7pt}
    \resizebox{\textwidth}{!}{
    \def\arraystretch{1.2}
    \begin{tabular} {l|c|c|c|c|c|c|c|c|c}
        \Xhline{2\arrayrulewidth}
        \multirow{3}{*}{Methods} & \multirow{3}{*}{Frames} & \multicolumn{8}{c}{NTU RGB+D 120 (\%)} \\
        \cline{3-10}
        & & \multicolumn{4}{c|}{X-Sub120} & \multicolumn{4}{c}{X-Set120} \\
        \cline{3-10}
        & & J & B & JM & BM & J & B & JM & BM \\
        \hline
        ST-GCN (*) \cite{STGCN} & 100 & 82.1 & 83.7 & 80.3 & 80.6 & 84.5 & 85.8 & 82.7 & 83.0 \\
        CTR-GCN \cite{CTRGCN} & 64 & 84.9 & 85.7 & 81.4 & 81.2 & - & 87.5 & - & - \\
        CTR-GCN (*) \cite{CTRGCN} & 100 & 84.0 & 85.9 & 81.1 & 82.2 & 85.9 & 87.4 & 84.1 & 83.9 \\
        ST-GCN++ \cite{STGCN++} & 100 & 83.2 & 85.6 & 80.4 & 81.5 & 85.6 & 87.5 & 84.3 & 83.0 \\
        InfoGCN \cite{InfoGCN} & 64 & 85.1 & 87.3 & 82.1 & \textbf{82.5} & 86.3 & 88.5 & 84.4 & \textbf{84.8} \\
        FR-Head \cite{FRHead} & 64 & 85.5 & 86.8 & 81.9 & 82.0 & 87.3 & 88.1 & 84.0 & 83.9 \\
        LST \cite{LST} & 64 & 85.5 & 87.5 & 82.3 & 82.4 & 87.0 & 88.7 & 83.9 & 84.4 \\
        HD-GCN \cite{HDGCN} & 64 & 85.7 & 86.7 & - & - & 87.3 & 88.4 & - & - \\
        \hline
        \cellcolor{c_lowbest}\textbf{SkateFormer} &\cellcolor{c_lowbest}64 &\cellcolor{c_lowbest}\textbf{87.7} &\cellcolor{c_lowbest}\textbf{88.2} &\cellcolor{c_lowbest}\textbf{83.1} &\cellcolor{c_lowbest}82.3 &\cellcolor{c_lowbest}\textbf{89.3} &\cellcolor{c_lowbest}\textbf{89.8} &\cellcolor{c_lowbest}\textbf{85.3} &\cellcolor{c_lowbest}84.1 \\
        \Xhline{2\arrayrulewidth}
    \end{tabular}}
    \label{tab:sinntu120}
\end{table*}

\subsection{Analysis of Partition-specific Attention}
\label{sec:b3}

\begin{figure}[htbp]
  \centering
  \includegraphics[width=1.0\textwidth]{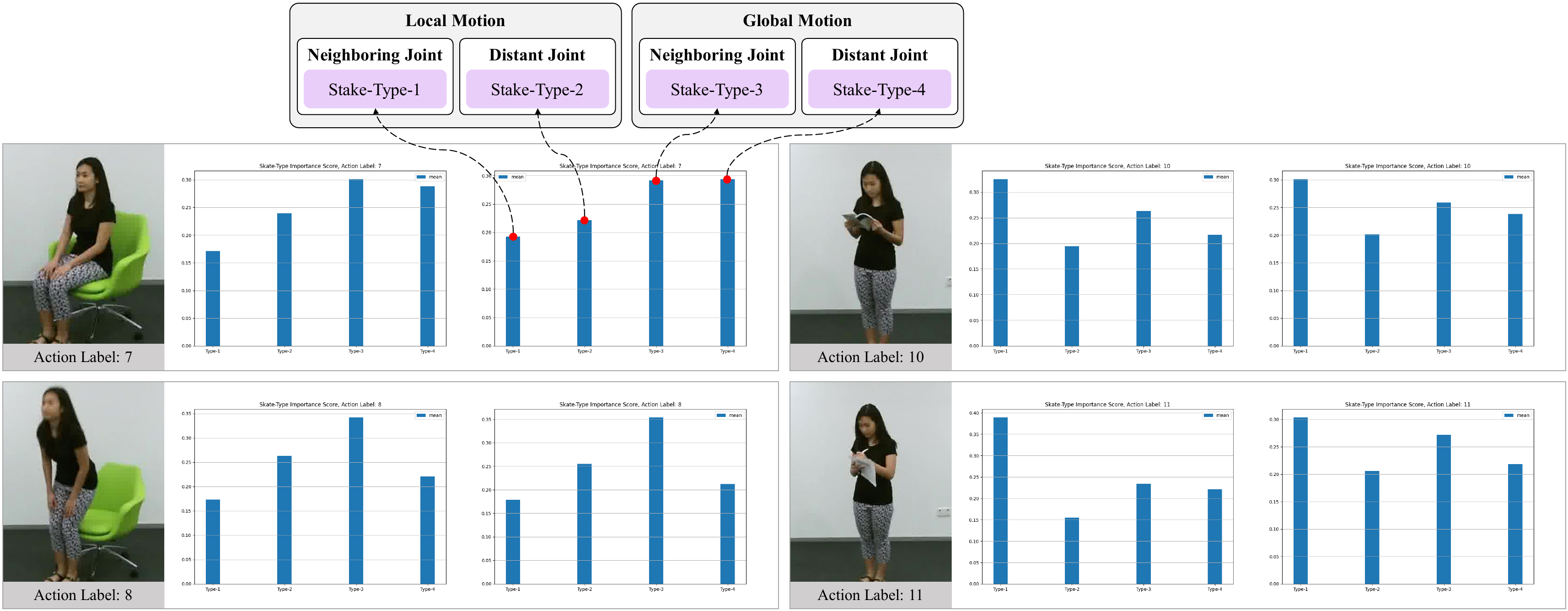}
  \caption{Skate-Type Importance Scores for various action labels.}
  \label{fig:score}
\end{figure}

To analyze the significance of different Skate-Types concerning action labels, we conducted an investigation within Skate-MSA. We assessed the strength of the feature correlation maps $\mathbf{Q}\mathbf{K}^{\mathsf{T}}$ (prior to $\mathsf{SoftMax}$ in $\mathsf{SA}$, Eq.~\ref{eq:8}) corresponding to each Skate-Type by calculating their mean values (referred to as \textit{Skate-Type Importance Score}) with respect to action labels. In Fig.~\ref{fig:score}, we illustrate the Skate-Type Importance Score for several action labels. For actions such as `sitting down' (action label: 7) and `standing up (from a sitting position)' (action label: 8), where the overall motion of the skeleton sequence is pivotal, the feature correlation maps for Skate-Type-3 or Skate-Type-4 exhibited more pronounced activations compared to other Skate-Types. Conversely, for actions that rely heavily on intricate hand movements like `reading' (action label: 10) and `writing' (action label: 11), the feature correlation map for Skate-Type-1 was prominently activated. Through this analysis, we affirm that our proposed partition-specific attention strategy based on Skate-Types operates adaptively according to action labels.

\begin{figure}
    \centering
    \includegraphics[width=1\textwidth]{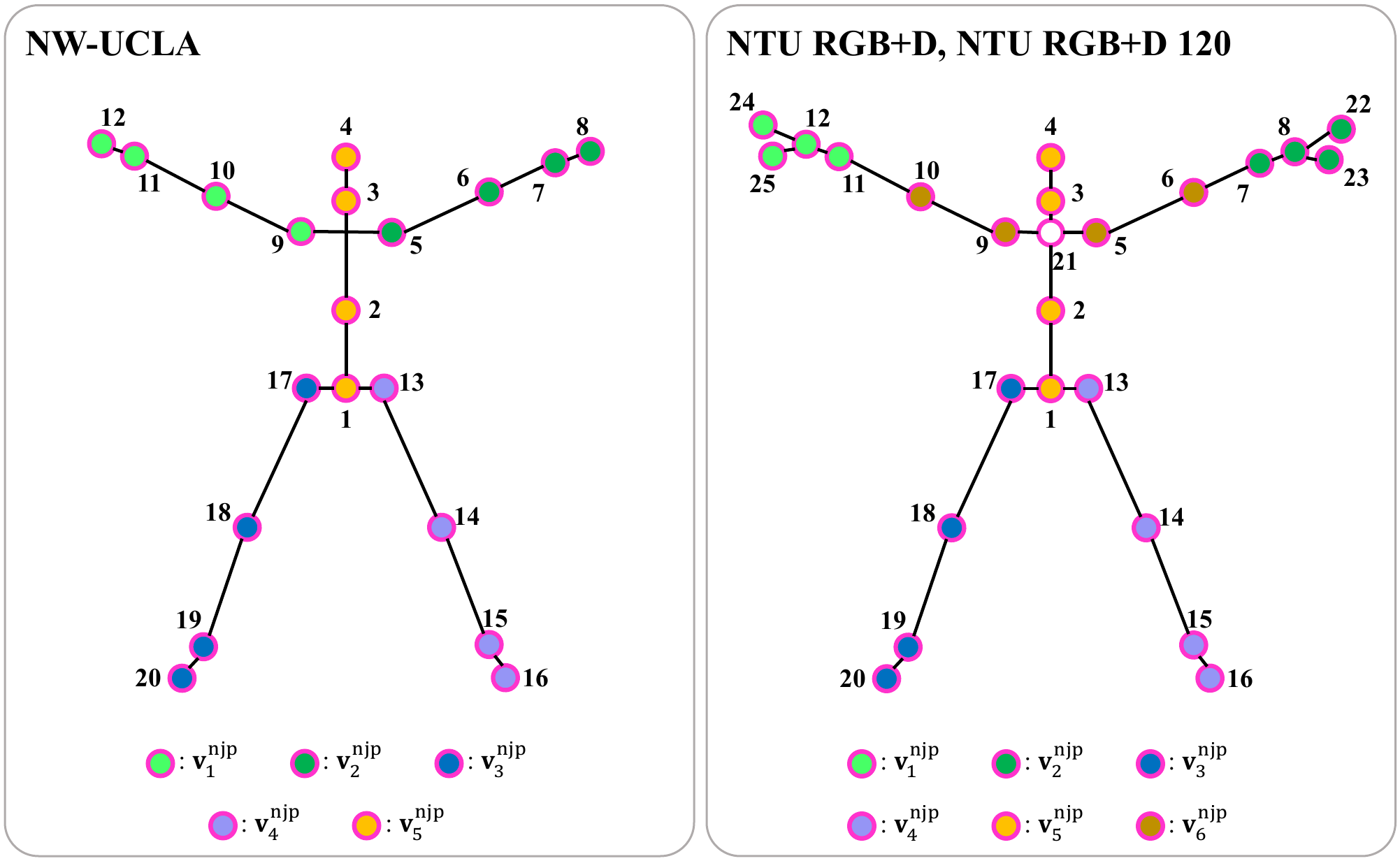}
    \captionsetup[figure]{hypcap=false}
    \captionof{figure}{Skeleton tracking indices and neighboring joint partitions.}
    \label{fig:joint}
\end{figure}

\section{Joint Details for Various Datasets}
\label{sec:joint}

\subsection{Skeleton Tracking Indices and Labels of Joints}

\noindent \textbf{NW-UCLA.}\quad In Fig.~\ref{fig:joint}, for the NW-UCLA \cite{NWUCLA} dataset acquired through the Kinect v1 \cite{Kinetics} sensor, the skeleton of a single individual consists of a total of 20 joints. Each joint, which is a constituent unit of the skeleton, can be represented by default indices from 1 to 20, as shown in Fig.~\ref{fig:joint}. The labels for each index are as follows \cite{Ntu120}: (1) base of spine, (2) middle of spine, (3) neck, (4) head, (5) left shoulder, (6) left elbow, (7) left wrist, (8) left hand, (9) right shoulder, (10) right elbow, (11) right wrist, (12) right hand, (13) left hip, (14) left knee, (15) left ankle, (16) left foot, (17) right hip, (18) right knee, (19) right ankle, (20) right foot. We will denote joints according to their tracking indices $k$ as $v_{k}^{\mathsf{tra}}$.

\noindent \textbf{NTU RGB+D and NTU RGB+D 120.}\quad In Fig.~\ref{fig:joint}, for the NTU RGB+D \cite{Ntu60} and NTU RGB+D 120 \cite{Ntu120} datasets acquired through the Kinect v2 \cite{Kinetics} sensor, the skeleton of a single individual consists of a total of 25 joints. These 25 joints include the original 20 joints from the Kinect v1 sensor, along with 5 additional joints representing the tips of both hands and the thumbs, as well as a spine joint. Similar to the NW-UCLA dataset, each joint can be represented by default indices from 1 to 25, as shown in Fig.~\ref{fig:joint}. The labels for each index are as follows \cite{Ntu120}: (1) base of spine, (2) middle of spine, (3) neck, (4) head, (5) left shoulder, (6) left elbow, (7) left wrist, (8) left hand, (9) right shoulder, (10) right elbow, (11) right wrist, (12) right hand, (13) left hip, (14) left knee, (15) left ankle, (16) left foot, (17) right hip, (18) right knee, (19) right ankle, (20) right foot, (21) spine, (22) tip of left hand, (23) left thumb, (24) tip of right hand, (25) right thumb.

\subsection{Details of Neighboring Joint Partitions}
In this paper, we introduce the concept of neighboring joint partitions ($\mathbf{v}_{k}^{\mathsf{njp}}$). We partitioned the entire set of joints into a total of $K$ non-overlapping subsets for our analysis. We empirically set $K$ proportional to the number of joints. We used the same number of joints for each partition since the implementation complexity was surged in usage of `reshaping' and `attention' functions due to the irregularity of input size. Alternative solutions as future work might be (i) reusing the same joints, (ii) trimming similar joints, or (iii) resizing the partition by linear interpolation between joints. The ordering of joints in the torso is not critical since their relative positions remain nearly constant (rigid) for any actions. We conducted experiments by varying the index order of the joints in the torso, which yielded little difference in performance. Below, we provide detailed information about the neighboring joint partitions for each dataset.

\noindent \textbf{NW-UCLA.}\quad For the NW-UCLA dataset, we set $K=5$. Therefore, the total number of joints, $V=20$, is divided into $L=20/5=4$ elements, creating neighboring joint partitions. Specifically, $\mathbf{v}_{1}^{\mathsf{njp}}$ represents the joints of the right arm, $\mathbf{v}_{2}^{\mathsf{njp}}$ represents the joints of the left arm, $\mathbf{v}_{3}^{\mathsf{njp}}$ represents the joints of the right leg, $\mathbf{v}_{4}^{\mathsf{njp}}$ represents the joints of the left leg, and $\mathbf{v}_{5}^{\mathsf{njp}}$ represents the vertical torso. These partitions are ordered in a manner that extends outward from the body's central region:

\begin{equation}
    \begin{split}
        \mathbf{v}_{1}^{\mathsf{njp}} &= \left[v_{9}^{\mathsf{tra}}, v_{10}^{\mathsf{tra}}, v_{11}^{\mathsf{tra}}, v_{12}^{\mathsf{tra}}\right]\\
        \mathbf{v}_{2}^{\mathsf{njp}} &= \left[v_{5}^{\mathsf{tra}}, v_{6}^{\mathsf{tra}}, v_{7}^{\mathsf{tra}}, v_{8}^{\mathsf{tra}}\right]\\
        \mathbf{v}_{3}^{\mathsf{njp}} &= \left[v_{17}^{\mathsf{tra}}, v_{18}^{\mathsf{tra}}, v_{19}^{\mathsf{tra}}, v_{20}^{\mathsf{tra}}\right]\\
        \mathbf{v}_{4}^{\mathsf{njp}} &= \left[v_{13}^{\mathsf{tra}}, v_{14}^{\mathsf{tra}}, v_{15}^{\mathsf{tra}}, v_{16}^{\mathsf{tra}}\right]\\
        \mathbf{v}_{5}^{\mathsf{njp}} &= \left[v_{2}^{\mathsf{tra}}, v_{3}^{\mathsf{tra}}, v_{1}^{\mathsf{tra}}, v_{4}^{\mathsf{tra}}\right].
    \end{split}
    \label{eq:nw}
\end{equation}

\noindent \textbf{NTU RGB+D and NTU RGB+D 120.}\quad For the NTU RGB+D and NTU RGB+D 120 datasets, we set $K=12$. Since each frame in the dataset may contain up to two individuals, and to ensure accurate action recognition, we treat each individual's joints separately. This results in a total of 50 joints per frame. In instances where only one individual is present, the remaining 25 joints are zero-padded. To facilitate partitioning, we exclude the 21st joint and consider the remaining 48 joints, resulting in $V=48$. Thus, we partition each individual's joints into $L=48/12=4$ elements for neighboring joint partitions: $\mathbf{v}_{1}^{\mathsf{njp}}$ for the right arm, $\mathbf{v}_{2}^{\mathsf{njp}}$ for the left arm, $\mathbf{v}_{3}^{\mathsf{njp}}$ for the right leg, $\mathbf{v}_{4}^{\mathsf{njp}}$ for the left leg, $\mathbf{v}_{5}^{\mathsf{njp}}$ for the vertical torso, and $\mathbf{v}_{6}^{\mathsf{njp}}$ for the horizontal torso. These partitions are ordered in a manner that extends outward from the body's central region:

\begin{equation}
    \begin{split}
        \mathbf{v}_{1}^{\mathsf{njp}} &= \left[v_{11}^{\mathsf{tra}}, v_{12}^{\mathsf{tra}}, v_{24}^{\mathsf{tra}}, v_{25}^{\mathsf{tra}}\right]\\
        \mathbf{v}_{2}^{\mathsf{njp}} &= \left[v_{7}^{\mathsf{tra}}, v_{8}^{\mathsf{tra}}, v_{22}^{\mathsf{tra}}, v_{23}^{\mathsf{tra}}\right]\\
        \mathbf{v}_{3}^{\mathsf{njp}} &= \left[v_{17}^{\mathsf{tra}}, v_{18}^{\mathsf{tra}}, v_{19}^{\mathsf{tra}}, v_{20}^{\mathsf{tra}}\right]\\
        \mathbf{v}_{4}^{\mathsf{njp}} &= \left[v_{13}^{\mathsf{tra}}, v_{14}^{\mathsf{tra}}, v_{15}^{\mathsf{tra}}, v_{16}^{\mathsf{tra}}\right]\\
        \mathbf{v}_{5}^{\mathsf{njp}} &= \left[v_{2}^{\mathsf{tra}}, v_{3}^{\mathsf{tra}}, v_{1}^{\mathsf{tra}}, v_{4}^{\mathsf{tra}}\right]\\
        \mathbf{v}_{6}^{\mathsf{njp}} &= \left[v_{5}^{\mathsf{tra}}, v_{9}^{\mathsf{tra}}, v_{6}^{\mathsf{tra}}, v_{10}^{\mathsf{tra}}\right].
    \end{split}
    \label{eq:ntu}
\end{equation}

\section{Details of Data augmentation}
\label{sec:detail} 

\subsection{Intra-instance Augmentation}

\noindent \textbf{Skeletal augmentation.}\quad In our work, we employed skeletal augmentation, which includes random shear (multiply matrix $\mathsf{Sh}$ for three coordinate axes), random rotation (multiply matrix $\mathsf{R}$ for two random coordinate axes), random scaling (multiply matrix $\mathsf{Sc}$ for three coordinate axes), random skeletal flipping (swap the indices of the joints for joints with left and right pairs), random coordinate dropout (randomly remove one coordinate axis), random joint dropout (randomly remove a subset of the entire set of joints), and actor order permutation (randomly rearrange the order of actors) \cite{AimCLR}. We did not utilize random temporal flip, random Gaussian noise, and random Gaussian blur \cite{ActCLR} as they were found to degrade the performance of SkateFormer. Detailed information regarding the skeletal augmentation techniques employed in our study can be found in Table~\ref{tab:skl}, where $p_{\mathsf{aug}}$ is application probability,

\begin{equation}
    \mathsf{Sh}=
    \begin{bmatrix}
        1 & \mathsf{sh}_{1} & \mathsf{sh}_{2} \\
        \mathsf{sh}_{1} & 1 & \mathsf{sh}_{2} \\
        \mathsf{sh}_{1} & \mathsf{sh}_{2} & 1 \\
    \end{bmatrix},
    \label{eq:sh}
\end{equation}

\begin{equation}
    \mathsf{R}=
    \begin{bmatrix}
        \cos\theta & -\sin\theta \\
        \sin\theta & \cos\theta \\
    \end{bmatrix},
    \label{eq:r}
\end{equation}

\noindent and

\begin{equation}
    \mathsf{Sc}=
    \begin{bmatrix}
        \mathsf{sc}_{1} & 0 & 0 \\
        0 & \mathsf{sc}_{2} & 0 \\
        0 & 0 & \mathsf{sc}_{3} \\
    \end{bmatrix}.
    \label{eq:sc}
\end{equation}

\begin{table}[th]
\scriptsize
    \centering
    \caption{Details of skeletal augmentation.}
    \setlength\tabcolsep{7pt}
    \resizebox{\textwidth}{!}{
    \def\arraystretch{1.2}
    \begin{tabular}{l|c|c}
        \Xhline{2\arrayrulewidth}
        \multirow{2}{*}{Skeletal Augmentation} & \multicolumn{2}{c}{Hyperparameters for Each Dataset} \\
        \cline{2-3}
        & NW-UCLA & NTU RGB+D, NTU RGB+D 120 \\
        \hline
        Random Shear & $p_{\mathsf{aug}}=0$ & $p_{\mathsf{aug}}=0.5$, $\mathsf{sh}_{i} \in [-0.5, 0.5]$ \\
        Random Rotation & $p_{\mathsf{aug}}=1$, $\theta \in [-\pi/3, \pi/3]$ & $p_{\mathsf{aug}}=0.5$, $\theta \in [-\pi/6, \pi/6]$ \\
        Random Scaling & $p_{\mathsf{aug}}=1$, $\mathsf{sc}_{i} \in [0.5, 1.5]$ & $p_{\mathsf{aug}}=0.5$, $\mathsf{sc}_{i} \in [0.8, 1.2]$ \\
        Random Skeletal Flipping & $p_{\mathsf{aug}}=0$ & $p_{\mathsf{aug}}=0.5$ \\
        Random Coordinate Dropout & $p_{\mathsf{aug}}=0$ & $p_{\mathsf{aug}}=0.5$ \\
        Random Joint Dropout & $p_{\mathsf{aug}}=0$ & $p_{\mathsf{aug}}=0.5$ \\
        Actor Order Permutation & $p_{\mathsf{aug}}=0$ & $p_{\mathsf{aug}}=0.5$ \\ 
        \Xhline{2\arrayrulewidth}
    \end{tabular}}
  \label{tab:skl}
\end{table}

\noindent \textbf{Temporal augmentation.}\quad Fig.~\ref{fig:5} illustrates our trimmed-uniform random sampling of frames with $p$ portion. For this, a $p \times 100 \; (0.5\leq p\leq 1)$ percentage of the total input sequence length ($T_{\mathsf{total}}$) is randomly determined before sampling. Given a $p$ value, a start-frame index $t_{s}$ is also randomly selected in the range of $0 \leq t_{s} \leq \lfloor T_{\mathsf{total}} \times (1-p) \rfloor$, and the end-frame index $t_{e}$ is determined as $t_{e} = t_{s} + \lfloor T_{\mathsf{total}} \times p \rfloor$, thus constituting the selected interval $[t_{s}, t_{e}]$ which is further uniformly divided into total $T$ sub-intervals. Then, as done in \cite{PoseC3D}, total $T$ frames are randomly selected with one random sampling in each sub-interval. When the selected interval $[t_{s}, t_{e}]$ is shorter than $T$ ($T=64$ in our case), we use linear interpolation to generate $T$ frames by a built-in PyTorch \cite{Pytorch} function \texttt{torch.nn.functional.interpolate}. In the training phase, we employed a random portion $p$ (where $0.5 \leq p \leq 1$) and a random start-frame index $t_{s}$ (where $0 \leq t_{s} \leq \lfloor T_{\mathsf{total}} \times (1-p) \rfloor$). However, during inference, for consistency in results, we set $p=0.95$ and $t_{s} = \lfloor T_{\mathsf{total}} \times (1-p) / 2 \rfloor$.

\noindent \textbf{Skate-Embedding.}\quad  Conventional sampling methods used in generating input skeleton sequences maintain either an absolute temporal gap (i.e., the fixed stride \cite{HDGCN, InfoGCN, FRHead}) or a relative temporal gap (i.e., uniform random \cite{STGCN++, PoseC3D}) between adjacent sampled frames. Therefore, embedding temporal indices, either absolutely or relatively, using learnable features has not posed a significant challenge. However, our sampling method relies on the random variable $p$ to determine sub-intervals for sampling. Consequently, even for sequences generated from the same instance, the temporal gap between adjacent sampled frames and the sampling index of the first frame can differ significantly. This presents limitations in conveying temporal index information to the network through learnable features.

When $\lfloor T_{\mathsf{total}} \times p \rfloor \geq T$, the sampled temporal indices are designated as $t_{\mathsf{idx}} = [t_{1}, t_{2}, ..., t_{T}]$, as illustrated in Fig~\ref{fig:5}. As mentioned, in cases of $\lfloor T_{\mathsf{total}} \times p \rfloor < T$, we perform linear interpolation across the entire index range $[t_{1}, t_{2}, ..., t_{\lfloor T_{\mathsf{total}} \times p \rfloor}]$ to create $t_{\mathsf{idx}} = [t_{1}^{\prime}, t_{2}^{\prime}, ..., t_{T}^{\prime}]$. The following is the detailed equation for our fixed temporal index features $\mathsf{TE}$ based on traditional sinusoidal positional embeddings \cite{Attention}:

\begin{equation}
    \begin{split}
    \mathsf{TE}[j, 2k] &= \sin(t_{\mathsf{idx}}[j]/10000^{2k/C})\\
    \mathsf{TE}[j, 2k+1] &= \cos(t_{\mathsf{idx}}[j]/10000^{2k/C}),
    \end{split}
    \label{eq:te}
\end{equation}
\noindent where $\mathsf{TE} \in \mathbb{R}^{T \times C}$.

\subsection{Inter-instance Augmentation}
We propose a novel inter-instance augmentation technique called \textit{Bone Length AdaIN}, drawing inspiration from Skeleton AdaIN \cite{ActCLR}. Bone Length AdaIN enhances diversity among subjects with varying body sizes by exchanging the bone lengths across different subjects in different frame sequences, rather than within each frame sequence. Detailed information on the methodology can be found in Algorithm~\ref{alg:adain}. We applied inter-instance augmentation with $p_{\mathsf{aug}}=0.2$.

\begin{algorithm}[t]
\caption{Bone Length AdaIN}\label{alg:adain}
\begin{algorithmic}[1]
\Statex \textbf{Input:} Input skeleton sequence $\mathbf{X} \in \mathbb{R}^{T \times V \times C_{\mathsf{in}}}$, (normalized) sampled temporal indices $\mathbf{t}_{\mathsf{idx}} \in [-1, 1]^{T \times 1 \times 1}$, input true label $\mathbf{y}$, entire set of skeleton sequences $\{\mathbf{X}^{n}\}_{n=1}^{N}$, and predefined adjacency matrix $\mathbf{A}$.
\Statex 
\Statex \textbf{Prepare reference sequence:}
\State Choose the reference skeleton sequence (before sampling) $\mathbf{X}_{\mathsf{f}}^{\mathsf{b.s.}} \in \mathbb{R}^{T_{\mathsf{total}, \mathsf{f}} \times V \times C_{\mathsf{in}}}$, which has same true label with $\mathbf{y}$ from $\{\mathbf{X}^{n}\}_{n=1}^{N}$.
\State Calculate the sampled temporal indices of $\mathbf{X}_{\mathsf{f}}^{\mathsf{b.s.}}$: $\mathbf{t}_{\mathsf{idx}, \mathsf{f}} = \mathsf{int}( (\mathbf{t}_{\mathsf{idx}} + 1)/2 \times T_{\mathsf{total}, \mathsf{f}}) \in \mathbb{N}^{T \times 1 \times 1}$.
\State Sample the $\mathbf{X}_{\mathsf{f}}^{\mathsf{b.s.}}$ based on $\mathbf{t}_{\mathsf{idx}, \mathsf{f}}$ (denoted as $\mathbf{X}_{\mathsf{f}} \in \mathbb{R}^{T \times V \times C_{\mathsf{in}}}$).
\Statex 
\Statex \textbf{Bone length calculation:}
\State Calculate the bone of each skeleton sequence $\mathbf{B}, \mathbf{B}_{\mathsf{f}} \in \mathbb{R}^{T \times V \times C_{\mathsf{in}}}$ from $\mathbf{X}, \mathbf{X}_{\mathsf{f}}$ based on $\mathbf{A}$.
\State Calculate the bone length of each skeleton sequence $\mathbf{L}, \mathbf{L}_{\mathsf{f}} \in \mathbb{R}^{T \times V \times 1}$: $\mathbf{L} = ||\mathbf{B}||_{2}$ and  $\mathbf{L}_{\mathsf{f}} = ||\mathbf{B}_\mathsf{f}||_{2}$.
\State Calculate the ratio of bone length: $\mathbf{r}=(\mathbf{L}_{\mathsf{f}} + \mathsf{eps})/(\mathbf{L} + \mathsf{eps}) \in \mathbb{R}^{T \times V \times 1}$, where $\mathsf{eps}=10^{-6}$.
\Statex 
\Statex \textbf{Bone length adaptation:}
\State Adapt the bone length of input sequence: $\mathbf{B}_{\mathsf{AdaIN}} = \mathbf{B} \times \mathbf{r}$.
\State Recover the skeleton sequence (joint) $\mathbf{X}_{\mathsf{AdaIN}}$ from $\mathbf{B}_{\mathsf{AdaIN}}$ based on $\mathbf{A}$.
\Statex 
\Statex \textbf{Output:} $\mathbf{X}_{\mathsf{AdaIN}}$
\end{algorithmic}
\end{algorithm}

\section{Self-Attention Analysis}
\label{sec:attn}

\subsection{Computational Complexity of Skate-MSA}
\label{sec:e1}

When given the feature maps $\mathbf{Q}$, $\mathbf{K}$, and $\mathbf{V}$ of feature map $\mathbf{x}_{\mathsf{msa}}$, the computational complexity of the naive self-attention layer is $2(VT)^2(C/2)$:

\begin{equation}
    \begin{split}
    \mathsf{Attn} = \mathsf{SoftMax}(\mathbf{Q}\mathbf{K}^{\mathsf{T}}) &\rightarrow [(VT)^{2}](\frac{C}{2}) \\
    \mathsf{SA}(\mathbf{x}_{\mathsf{msa}}) = \mathsf{Attn}\cdot\mathbf{V}  &\rightarrow [(VT)\frac{C}{2}](VT),
    \end{split}
    \label{eq:naive}
\end{equation}

\noindent where $\mathbf{Q}, \mathbf{K}, \mathbf{V} \in \mathbb{R}^{(T \times V) \times C/2}$.

Let's calculate the computation complexity for our proposed Skate-MSA. $\mathbf{x}_{\mathsf{msa}}$ is first split channel-wise ($\mathbf{x}_{\mathsf{msa}}^{i}$) and then partitioned into $\mathbf{x}_{\mathsf{msa}}^{i,\mathcal{P}}$. When considering the corresponding $\mathbf{Q}$, $\mathbf{K}$, and $\mathbf{V}$ for $\mathbf{x}_{\mathsf{msa}}^{i,\mathcal{P}}$, the computation complexity becomes $B \cdot 2(V^{\prime}T^{\prime})^2(C/8)$, where $\mathbf{Q}, \mathbf{K}, \mathbf{V} \in \mathbb{R}^{B \times (T^{\prime} \times V^{\prime}) \times C/8}$. This can be easily computed in a manner similar to Eq.~\ref{eq:naive}. The computation complexity for each Skate-Type is as follows:

\begin{equation}
    \begin{split}
    \text{Skate-Type-1:}& (MK) \cdot 2(LN)^2\frac{C}{8}=2(VT)^2\frac{C}{2}\left[\frac{1}{4}\cdot\frac{1}{MK}\right] \\
    \text{Skate-Type-2:}& (ML) \cdot 2(KN)^2\frac{C}{8}=2(VT)^2\frac{C}{2}\left[\frac{1}{4}\cdot\frac{1}{ML}\right] \\
    \text{Skate-Type-3:}& (NK) \cdot 2(LM)^2\frac{C}{8}=2(VT)^2\frac{C}{2}\left[\frac{1}{4}\cdot\frac{1}{NK}\right] \\
    \text{Skate-Type-4:}& (NL) \cdot 2(KM)^2\frac{C}{8}=2(VT)^2\frac{C}{2}\left[\frac{1}{4}\cdot\frac{1}{NL}\right],
    \end{split}
    \label{eq:skate}
\end{equation}

\noindent where $V=KL$ and $T=MN$. Therefore, by summing up all the complexities in Eq.~\ref{eq:skate}, the overall complexity of Skate-MSA is as follows:

\begin{equation}
    2(VT)^{2}\frac{C}{2}\left[\frac{1}{4}\left(\frac{1}{MK}+\frac{1}{ML}+\frac{1}{NK}+\frac{1}{NL}\right)\right].
    \label{eq:total}
\end{equation}

\subsection{Additional Experiments on the Computational Complexity}
\label{sec:e2}

Our SkateFormer comprises 8 SkateFormer blocks and 3 temporal downsampling layers. The 48$\times$ reduction in FLOPs was \textit{theoretically calculated} from the first two SkateFormer blocks before downsampling, where the reduction effect is most pronounced. To clarify this, we additionally performed experiments in Table~\ref{tab:e2} in terms of measured FLOPs and memory usages in perspective of the attention layers \textit{only}. As shown, the Skate-MSA is 38.87$\times$ less complex in terms of FLOPs for the first two attention layers.

\begin{table}[htp]
\scriptsize
    \centering
    \caption{FLOPs and memory usage for attention layers.}
    \setlength\tabcolsep{7pt}
    \resizebox{0.99\textwidth}{!}{
    \def\arraystretch{1.2}
    \begin{tabular} {l|c|c|c|c}
        \Xhline{2\arrayrulewidth}
        \multirow{2}{*}{\makecell[l]{Attention Types\\in attention layer}} & \multicolumn{2}{c|}{Total 8 attention layers} & \multicolumn{2}{c}{\cellcolor{c_lowbest}First two attention layers} \\
        \cline{2-5}
        & FLOPs (M) & Memory (MB) & FLOPs (M) & Memory (MB) \\
        \hline
        $\mathbf{v}_{k}^{\mathsf{njp}}$ + $\mathbf{v}_{l}^{\mathsf{djp}}$ + $\mathbf{t}_{m}^{\mathsf{local}}$ + $\mathbf{t}_{n}^{\mathsf{global}}$ & 29.12 & 56.3 & 15.53 & 27.0 \\
        \hline
        Naive self-attention & 801.79 (\textcolor{red}{$\times$27.53}) & 1530.0  (\textcolor{red}{$\times$27.18}) & \cellcolor{c_lowbest}603.65  (\textbf{\textcolor{red}{$\times$38.87}}) & \cellcolor{c_lowbest}1152.0  (\textbf{\textcolor{red}{$\times$42.67}}) \\
        \Xhline{2\arrayrulewidth}
    \end{tabular}}
    \label{tab:e2}
\end{table}

\section{Details of Implementation}

\subsection{Hyperparameters}
Here is a more detailed description of the hyperparameters we used. We employed the AdamW optimizer \cite{AdamW} with $\beta_{1}=0.9$, $\beta_{2}=0.999$, and a weight decay value of 0.1. For the cosine-annealing scheduler \cite{CosineAnneal}, we set warmup learning rate to $10^{-7}$, minimum learning rate to $10^{-5}$, base learning rate to $10^{-3}$, warm-up epochs to 25, and utilized linear warm-up strategy. To prevent overfitting in the network, we set the attention map dropout ratio to 0.5 and the drop path ratio to 0.2. We did not use dropout for classification head and linear layers. The number of heads $H$ was consistent across all SkateFormer Blocks, set at 32. The expansion ratio of FFN was set to 1 for the NW-UCLA \cite{NWUCLA} dataset and 4 for the NTU RGB+D \cite{Ntu60} and NTU RGB+D 120 \cite{Ntu120} datasets. We employed the GELU activation function in our model. Furthermore, we employed a total of $R=8$ SkateFormer Blocks, applying temporal downsampling after every 2 blocks. The kernel size for the $\mathsf{T{\text -}Conv}$ layers was set to $k=7$. For the performance evaluation, we compare our SkateFormer with other methods in terms of the top-1 action recognition accuracy for the test data sets.

\subsection{Overall Architecture}
The overall architecture of SkateFormer and the output shape of feature maps produced by each module are presented in Table~\ref{tab:overall}.

\begin{table}[htp]
\scriptsize
    \centering
    \caption{The details of SkateFormer.}
    \setlength\tabcolsep{7pt}
    \resizebox{0.6\textwidth}{!}{
    \def\arraystretch{1.2}
    \begin{tabular} {c|c|c}
        \Xhline{2\arrayrulewidth}
        & Mudule & Output Shape \\
        \hline
        Input & & $64 \times V \times 3$ \\
        \hline
        \multirow{3}{*}{Projection} & $\mathsf{Linear}$ 1 & $64 \times V \times 6$ \\
        & $\mathsf{Linear}$ 2 & $64 \times V \times 9$ \\
        & $\mathsf{Linear}$ 3 & $64 \times V \times 96$ \\
        \hline
        \multirow{2}{*}{Stage 1} & SkateFormer Block 1 & $64 \times V \times 96$ \\
        & SkateFormer Block 2 & $64 \times V \times 96$ \\
        \hline
        Downsampling & $\mathsf{T{\text -}Conv} \; (\downarrow_{2})$ 1 & $32 \times V \times 96$ \\
        \hline
        \multirow{2}{*}{Stage 2} & SkateFormer Block 3 & $32 \times V \times 192$ \\
        & SkateFormer Block 4 & $32 \times V \times 192$ \\
        \hline
        Downsampling & $\mathsf{T{\text -}Conv} \; (\downarrow_{2})$ 2 & $16 \times V \times 192$ \\
        \hline
        \multirow{2}{*}{Stage 3} & SkateFormer Block 5 & $16 \times V \times 192$ \\
        & SkateFormer Block 6 & $16 \times V \times 192$ \\
        \hline
        Downsampling & $\mathsf{T{\text -}Conv} \; (\downarrow_{2})$ 3 & $8 \times V \times 192$ \\
        \hline
        \multirow{2}{*}{Stage 4} & SkateFormer Block 7 & $8 \times V \times 192$ \\
        & SkateFormer Block 8 & $8 \times V \times 192$ \\
        \hline
        \multirow{2}{*}{Head} & Pool & $1 \times 1 \times 192$ \\
        & $\mathsf{Linear}$ (classification) & $1 \times 1 \times N_{c}$ \\
        \hline
        Output & & $1 \times 1 \times N_{c}$ \\
        \Xhline{2\arrayrulewidth}
    \end{tabular}}
  \label{tab:overall}
\end{table}

\subsection{Modules}

\noindent \textbf{Details of $\mathsf{G{\text -}Conv}$.}\quad Let us denote $\mathbf{G}$ as learnable parameters, where $\mathbf{G} \in \mathbb{R}^{V \times V \times H/4}$. The input to $\mathsf{G{\text -}Conv}$, denoted as $\mathbf{x}_\mathsf{gc}$, is split into $H/4 (=32/4=8)$ features in a channel-wise manner:

\begin{equation}
    [\mathbf{x}_{\mathsf{gc}}^{1}, ..., \mathbf{x}_{\mathsf{gc}}^{H/4}] = \mathsf{Split}(\mathbf{x}_{\mathsf{gc}}),
    \label{eq:splitgc}
\end{equation}

\noindent where $\mathbf{x}_\mathsf{gc} \in \mathbb{R}^{T \times V \times C/4}$. Each $\mathbf{x}_{\mathsf{gc}}^{h}$ undergoes matrix multiplication with its corresponding $\mathbf{G}_{h} = \mathbf{G}[:, :, h]$, and the results are channel-wise concatenated to form the output of $\mathsf{G{\text -}Conv}$:

\begin{equation}
    \mathsf{G{\text -}Conv}(\mathbf{x}_{\mathsf{gc}}) = \mathsf{Concat}(\mathbf{G}_{1}\cdot\mathbf{x}_{\mathsf{gc}}^{1}, ..., \mathbf{G}_{H/4}\cdot\mathbf{x}_{\mathsf{gc}}^{H/4}),
    \label{eq:splitgc2}
\end{equation}

\noindent where $\mathbf{G}_{h} \in \mathbb{R}^{V \times V}$. Through this process, we obtain an inductive bias with respect to skeletal position.

\noindent \textbf{Details of FFN.}\quad The FFN of the SkateFormer Block is expressed as follows: $\mathbf{x} \leftarrow \mathbf{x} + \mathsf{Linear}(\mathsf{Act}(\mathsf{Linear}(\mathsf{LN}(\mathbf{x}))))$, where $\mathbf{x} \in \mathbb{R}^{T \times V \times C}$. If we denote the expansion ratio as $e$, the first linear layer in the FFN expands features from $C$ channels to a higher-dimensional $e \times C$ channels, and the second linear layer squeezes them back to $C$ channels from $e \times C$ channels. For the NW-UCLA dataset, we set $e=1$, while for the NTU RGB+D and NTU RGB+D 120 datasets, we used $e=4$.

\begin{table*}[ht]
\scriptsize
    \centering
    \caption{List of symbols.}
    \setlength\tabcolsep{7pt}
    \resizebox{\textwidth}{!}{
    \def\arraystretch{1.3}
    \begin{tabular}{l|c|l}
        \Xhline{2\arrayrulewidth}
        & Symbol & Description \\
        \hline
        \multirow{5}{*}{Constant} & $T_{\mathsf{total}}$ & The total input sequence length (Varies from one instance to another) \\
        & $T$ & The number of sampled frames ($T=64$ in our case) \\
        & $V$ & The number of total joints \\
        & $C_{\mathsf{in}}$ & The dimensionality of data representing a single joint ($C_{\mathsf{in}}=3$ in our case) \\
        & $N_{c}$ & The number of classes \\
        \hline
        \multirow{7}{*}{\makecell{Hyper-\\parameter}} & $C$ & The dimensionality of feature map \\
        & $H$ & The number of total heads \\
        & $R$ & The number of total SkateFormer Blocks, $R_{1}+R_{2}+R_{3}+R_{4}$ \\
        & $K$ & The number of neighboring joint partitions \\
        & $L$ & The number of distant joint partitions \\
        & $M$ & The number of local frame partitions \\
        & $N$ & The number of global frame partitions \\
        \hline
        \multirow{3}{*}{Input} & $\mathbf{X}$ & Input sampled skeleton sequence ($\mathbf{X} \in \mathbb{R}^{T \times V \times C_{\mathsf{in}}}$) \\
        & $\mathbf{f}^{t}$ & $t$-th frame of input skeleton sequence ($\mathbf{f}^{t} \in \mathbb{R}^{V \times C_{\mathsf{in}}}$) \\
        & $t_{\mathsf{idx}}$ & The sampled temporal indices ($t_{\mathsf{idx}} \in \mathbb{R}^{T}$) \\
        \hline
        \multirow{2}{*}{Output} & $\hat{\mathbf{y}}$ & The outcome of SkateFormer ($\hat{\mathbf{y}} \in \mathbb{R}^{N_{c}}$) \\
        & $\mathbf{y}$ & The true label ($\mathbf{y} \in \mathbb{R}^{N_{c}}$) \\
        \hline
        \multirow{8}{*}{Skate-MSA} & $\bm{\mathcal{P}}_{i}$ & The $i$-th Skate-Type partition operation \\
        & $\bm{\mathcal{R}}_{i}$ & The $i$-th Skate-Type reverse operation \\
        & $\mathbf{v}$ & The skeletal axis of feature map $\mathbf{x}$ \\
        & $\mathbf{t}$ & The time axis of feature map $\mathbf{x}$ \\
        & $\mathbf{v}_{k}^{\mathsf{njp}}$ & The $k$-th neighboring joint partition, $\mathbf{v} = \{\mathbf{v}_{k}^{\mathsf{njp}}\}_{k=1}^{K}$ \\
        & $\mathbf{v}_{l}^{\mathsf{djp}}$ & The $l$-th distant joint partition, $\mathbf{v} = \{\mathbf{v}_{l}^{\mathsf{djp}}\}_{l=1}^{L}$ \\
        & $\mathbf{t}_{m}^{\mathsf{local}}$ & The $m$-th local frame partition, $\mathbf{t} = \{\mathbf{t}_{m}^{\mathsf{local}}\}_{m=1}^{M}$\\
        & $\mathbf{t}_{n}^{\mathsf{global}}$ & The $n$-th global frame partition, $\mathbf{t} = \{\mathbf{t}_{n}^{\mathsf{global}}\}_{n=1}^{N}$ \\
        \hline
        \multirow{6}{*}{Feature} & $\mathbf{x}$ & The input for SkateFormer Block ($\mathbf{x} \in \mathbb{R}^{T \times V \times C}$) \\
        & $\mathbf{x}_{\mathsf{gc}}$ & The input for $\mathsf{G{\text -}Conv}$ ($\mathbf{x}_{\mathsf{gc}} \in \mathbb{R}^{T \times V \times \frac{C}{4}}$) \\
        & $\mathbf{x}_{\mathsf{tc}}$ & The input for $\mathsf{T{\text -}Conv}$ ($\mathbf{x}_{\mathsf{tc}} \in \mathbb{R}^{T \times V \times \frac{C}{4}}$) \\
        & $\mathbf{x}_{\mathsf{msa}}$ & The input for Skate-MSA ($\mathbf{x}_{\mathsf{msa}} \in \mathbb{R}^{T \times V \times \frac{C}{2}}$) \\
        & $\mathbf{x}_{\mathsf{msa}}^{i}$ & The input for MSA of $i$-th Skate-Type ($\mathbf{x}_{\mathsf{msa}}^{i} \in \mathbb{R}^{T \times V \times \frac{C}{8}}$) \\
        & $\mathbf{x}_{\mathsf{msa}}^{i, \bm{\mathcal{P}}}$ & The partitioned $\mathbf{x}_{\mathsf{msa}}^{i}$, $\bm{\mathcal{P}}_{i}(\mathbf{x}_{\mathsf{msa}}^{i})$\\
        \Xhline{2\arrayrulewidth}
    \end{tabular}}
  \label{tab:sym}
\end{table*}

\end{sloppypar}
\end{document}